\documentclass[10pt,twocolumn,letterpaper]{article}

\usepackage{iccv}              % To produce the CAMERA-READY version
\usepackage{tikz}
\usepackage{amsmath,amssymb,mathtools}
\usepackage{algorithm}
\usepackage{algpseudocode}
\usepackage{multirow}
\usepackage{booktabs}
\definecolor{iccvblue}{rgb}{0.21,0.49,0.74}
\usepackage[pagebackref,breaklinks,colorlinks,allcolors=iccvblue]{hyperref}

\usepackage{colortbl}

\definecolor{tabfirst}{rgb}{1, 0.7, 0.7} % red
\definecolor{tabsecond}{rgb}{1, 0.85, 0.7} % orange
\definecolor{tabthird}{rgb}{1, 1, 0.7} % yellow
\newcommand{\fr}{{\cellcolor{tabfirst}}}
\newcommand{\nd}{{\cellcolor{tabsecond}}}

\title{Visibility-Aware Language Aggregation\\for Open-Vocabulary Segmentation in 3D Gaussian Splatting}

\author{
    Sen Wang$^{1,2}$ \quad
    Kunyi Li$^{1,2}$ \quad
    Siyun Liang$^{1, 5}$\quad
    Elena Alegret $^{1}$ \quad 
    Jing Ma $^{4}$ \quad \\
    Nassir Navab$^{1,2}$ \quad
    Stefano Gasperini$^{1,2,3}$ \\ 
    {\normalsize $^1$Technical University of Munich} \quad
    {\normalsize $^2$Munich Cental for Machine Learning} \quad
    {\normalsize $^3$VisualAIs} \\
    {\normalsize $^4$Ludwig Maximilian University of Munich} \quad
    {\normalsize $^5$University of Tübingen} \quad\\
    }
    
\begin{document}
\maketitle
\begin{abstract}
Recently, distilling open-vocabulary language features from 2D images into 3D Gaussians has attracted significant attention. Although existing methods achieve impressive language-based interactions with 3D scenes, we observe two fundamental issues: background Gaussians, which contribute negligibly to a rendered pixel, receive the same feature as the dominant foreground ones, and multi-view inconsistencies due to view-specific noise in language embeddings. We introduce Visibility-Aware Language Aggregation (VALA), a lightweight yet effective method that computes marginal contributions for each ray and applies a visibility-aware gate to retain only visible Gaussians. Moreover, we propose a streaming weighted geometric median in cosine space to merge noisy multi-view features. Our method yields a robust, view-consistent language feature embedding in a fast and memory-efficient manner. VALA improves open-vocabulary localization and segmentation across reference datasets, consistently surpassing existing works. The source code is available on \href{https://github.com/changandao/VALA}{VALA}.
\end{abstract}    
\section{Introduction}
\label{sec:intro}
% 3D scene understanding
Understanding 3D scenes is essential for interacting with the environment in robotic navigation~\cite{cadena2016slam, mur_artal_2017_orbslam2}, autonomous driving~\cite{geiger2012kitti, sun2020waymo}, and augmented reality~\cite{klein2007ptam, izadi2011kinectfusion}. Traditional approaches, however, are constrained to a fixed set of object categories defined at training time~\cite{qi2017pointnet, choy2019minkowski, thomas2019kpconv}, limiting their applicability to open-world scenarios.
% open-vocabulary
Thanks to recent advances in vision-language models~\cite{radford2021clip, jia2021align}, open-vocabulary methods~\cite{vild2021, detic2022, openscene2023} enable querying and interacting with 3D scenes through natural language, and recognizing unseen object categories without requiring retraining.

While classical 3D understanding methods operate on point clouds or meshes derived from 3D sensors, recent neural scene representations, such as NeRFs~\cite{mildenhall2020nerf} and 3D Gaussian Splatting (3DGS)~\cite{kerbl20233d}, have emerged as a compelling alternative. They not only enable high-quality rendering from novel viewpoints but also facilitate semantic reasoning, as appearance and geometry are encoded jointly. Thus, open-vocabulary reasoning has recently been grounded in neural 3D scene representations~\cite{kerr2023lerf, qin2024langsplat}, enabling new semantic interactions in 3D environments.
Initially explored with NeRFs~\cite{kerr2023lerf,engelmann2024opennerf}, the efficiency and explicit nature of 3DGS simplified the integration of semantic features, contributing to its widespread adoption~\cite{qin2024langsplat,wu2024opengaussian,drsplat25,cheng2024occam}.

\begin{figure}[t]
    \centering
   \includegraphics[width=\linewidth]{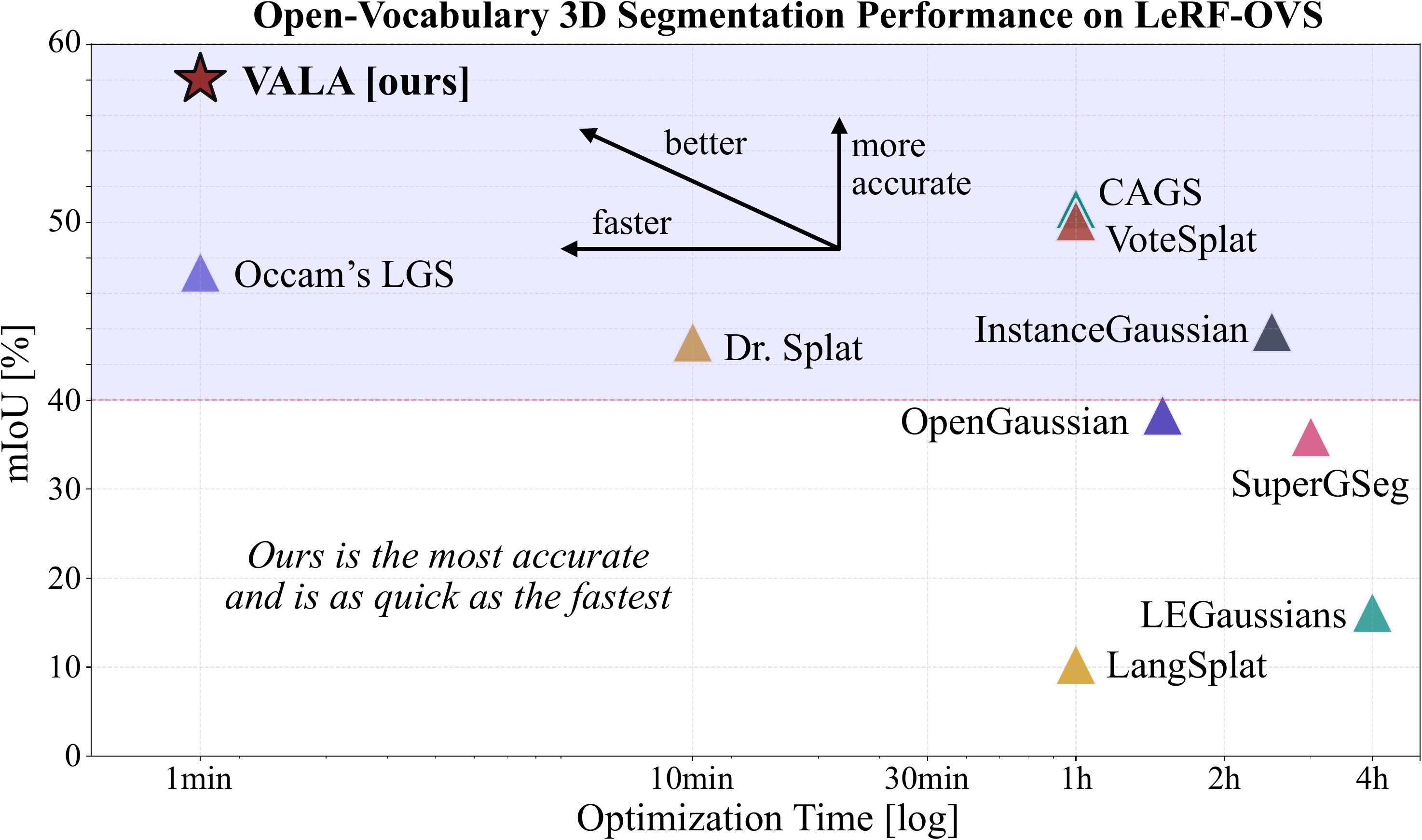}
   \vspace{-5mm}
    \caption{Thanks to its feature aggregation that is visibility-aware and multi-view consistent, our proposed VALA is the most accurate and as quick as the fastest~\cite{cheng2024occam} to optimize. Comparison in 3D open-vocabulary segmentation on the LeRF-OVS dataset~\cite{qin2024langsplat}.}
    \vspace{-5mm}
    \label{fig:teaser}
\end{figure}

At the core of these approaches lies the challenge of embedding reliable semantic and language features into the 3D representation. Current methods rely on powerful off-the-shelf 2D foundation models, such as SAM~\cite{kirillov2023sam} and CLIP~\cite{radford2021clip}, which produce 2D feature maps that must be lifted to 3D and aggregated across views. Proper aggregation is critical for accurate 3D segmentation.

Despite numerous recent advances~\cite{drsplat25,jiang2025votesplat,li2024instancegaussian,sun2025cags}, current approaches suffer from an inherent limitation: they assign 2D features indiscriminately to \textit{all} Gaussians along a camera ray, disregarding scene geometry and occlusion relationships. Consequently, features originating from foreground objects (\eg, a vase) are incorrectly propagated to background structures (\eg, the supporting table or floor), leading to substantial degradation in open-vocabulary recognition accuracy.

Furthermore, when lifted into 3D, 2D features exhibit multi-view inconsistencies. The same object may produce divergent feature representations across different viewpoints, a phenomenon known as semantic drift~\cite{kerr2023lerf}. Current methods address this by promoting cross-view consistency through 3D-consistent clustering and contrastive objectives derived from SAM masks~\cite{wu2024opengaussian, liang2024supergseg, li2024instancegaussian, piekenbrinck2025opensplat3d}. Nevertheless, such strategies generally require extensive per-scene optimization, and their heavy reliance on noisy, view-dependent 2D cues often undermines cluster reliability.

In this paper, we address these fundamental feature aggregation problems with VALA (Visibility-Aware Language Aggregation), a lightweight yet effective framework that combines a two-stage gating mechanism with a robust multi-view feature aggregation strategy. Our gating mechanism leverages the statistical distribution of per-ray Gaussian contributions (termed visibility) to preferentially propagate features to Gaussians with high visibility, thereby ensuring accurate feature assignment. To further mitigate multi-view inconsistencies in 2D language features, we introduce a convex but non-smooth optimization on the unit hypersphere, which we reformulate into a streaming gradient-based procedure that achieves consistent embeddings without additional computational overhead. As shown in Figure~\ref{fig:teaser}, VALA strategies are highly effective.

Our contributions can be summarized as follows:
\begin{itemize}
    \item We identify fundamental issues in the feature aggregation of current works as a bottleneck in open-vocabulary 3D scene understanding.
    \item We introduce VALA, a visibility-aware feature propagation framework that employs a two-stage gating mechanism to assign features based on Gaussian visibility.
    % to tackle this problem with a visibility-aware feature propagation leveraging a two-stage gating mechanism.
    \item We propose a robust aggregation strategy for the 2D features using the streaming cosine median, thereby improving multi-view consistency.
    \item We obtain state-of-the-art performance in 2D \textit{and} 3D on open-vocabulary segmentation for 3DGS scenes on the reference datasets LeRF-OVS~\cite{qin2024langsplat} and ScanNet-v2~\cite{dai2017scannet}.
\end{itemize}

\section{Related works}
\label{sec:releated_works}

\textbf{Open-Vocabulary Feature Distillation.}
Recent works have embedded 2D vision-language features into 3D scene representations to enable open-vocabulary 3D understanding. Pioneering efforts on NeRFs such as LERF~\cite{kerr2023lerf} and OpenNeRF~\cite{engelmann2024opennerf} used CLIP~\cite{radford2021clip} embeddings and pixel-aligned features, enabling open-vocabulary queries.
However, due to the computational needs of NeRF~\cite{mildenhall2020nerf}, they face scalability and efficiency bottlenecks. Thus, subsequent works have embedded language features into 3DGS~\cite{zuo2024fmgs, shi2024language, zhou2024feature}. LangSplat~\cite{qin2024langsplat} employs SAM~\cite{kirillov2023sam} to extract multi-level CLIP features, then compresses dimensionality with an autoencoder to build a compact yet expressive 3D language field. Feature3DGS~\cite{zhou2024feature} uses a convolutional neural network (CNN) to lift feature dimensions. Although both approaches aim to compress the supervision signal, this dimensionality reduction inevitably results in information loss. GOI~\cite{goi2024} and CCL-LGS~\cite{tian2025ccl} employ a single trainable feature codebook to store language embeddings, with an MLP predicting discrete codebook indices for rasterized 2D feature maps, which compress semantics spatially rather than dimensionally and retain semantic richness. However, as these approaches rely on 2D rendered feature maps for perception, their performance in 3D scene understanding is significantly limited. 

Other methods first group 3D Gaussians or points into semantically meaningful clusters, typically corresponding to objects or parts, and then assign a language feature to each cluster as a whole~\cite{liang2024supergseg, wu2024opengaussian, goi2024, li2024instancegaussian, piekenbrinck2025opensplat3d, jiang2025votesplat}. These methods introduce an explicit discrete grouping step as a form of prior semantic structuring: OpenGaussian~\cite{wu2024opengaussian} performs coarse-to-fine clustering based on spatial proximity followed by feature similarity. SuperGSeg~\cite{liang2024supergseg} and InstanceGaussian~\cite{li2024instancegaussian} both leverage neural Gaussians to model instance-level features: SuperGSeg groups Gaussians into Super-Gaussians to facilitate language assignment, whereas InstanceGaussian directly assigns fused semantic features to each cluster. VoteSplat~\cite{jiang2025votesplat} and OpenSplat3D~\cite{piekenbrinck2025opensplat3d} mitigate the pixel-level ambiguities of the direct distillation. Then, the resulting cluster graph structures support higher-level reasoning~\cite{liang2024supergseg, zhan2025hi}, which per-Gaussian features cannot easily enable. However, all these methods rely on feature distillation using per-cluster learnable language embeddings. These approaches are computationally expensive and highly sensitive to noise or outliers in the preprocessed feature maps, since the language features are optimized directly in Euclidean space. As a result, even minor errors in the input features can propagate through the model, leading to inconsistent or inaccurate semantic representations, particularly in complex or cluttered scenes.

\begin{figure*}[t]
    \centering
    \includegraphics[width=\linewidth]{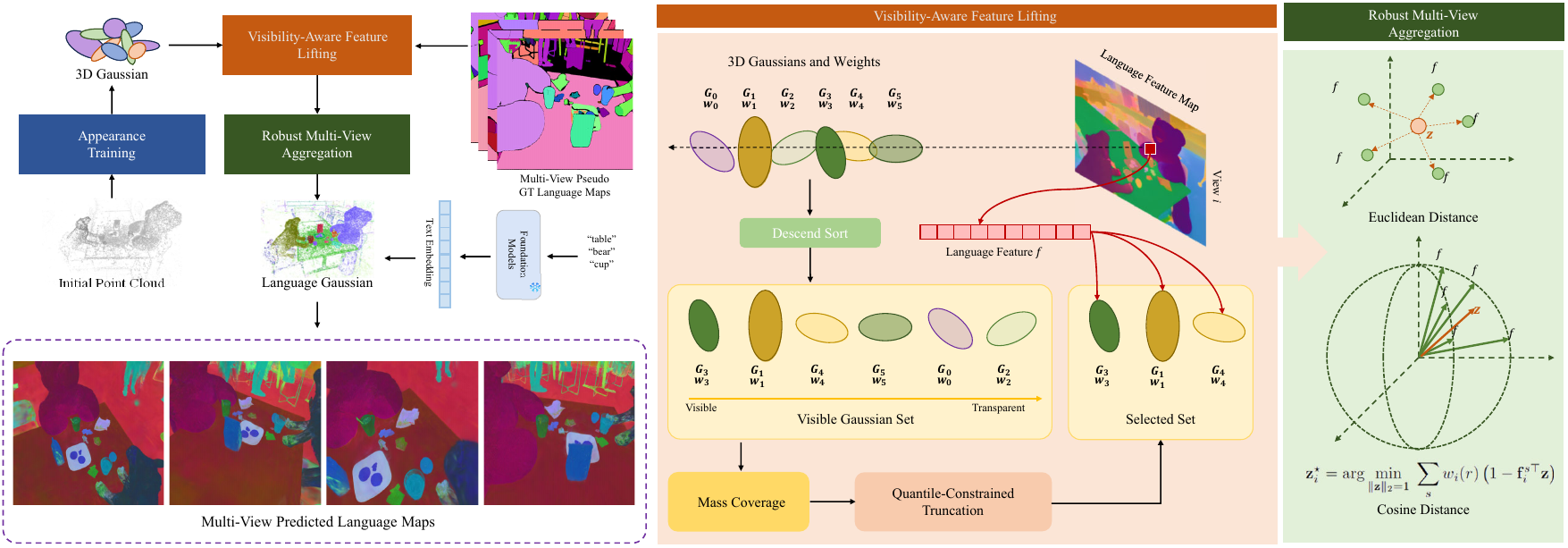}
    \vspace{-5mm}
    \caption{Overview of VALA. The framework is shown on the left, with the orange and green blocks detailed on the right being our key contributions: the visibility-aware feature lifting (orange, Section~\ref{sec:gating}), and the robust multi-view aggregation (green, Section~\ref{sec:median}).}
    \label{fig:main_method}
    \vspace{-5mm}
\end{figure*}

\textbf{Open-Vocabulary Feature Aggregation.}
Beyond cluster-based language features distillation, recent works adopt more efficient strategies for feature aggregation. For instance, Dr.Splat~\cite{drsplat25} and Occam's LGS~\cite{cheng2024occam} bypass intermediate 2D supervision and clustering by directly injecting language features into 3D Gaussians, achieving fast, accurate results in a training-free regime. While these direct feature aggregation methods deliver strong runtime efficiency and segmentation accuracy, they indiscriminately propagate 2D features to \emph{every} Gaussian intersected by each camera ray, disregarding scene geometry and occlusion. As a result, features from foreground objects (e.g., a vase) are erroneously assigned to background elements (e.g., the table or floor). Moreover, existing methods share two critical limitations: (i) they assign equal supervision to all Gaussians along a ray, ignoring each Gaussian’s marginal contribution to the rendered pixel, and (ii) they overlook the view-dependent noise and inconsistency in 2D language features. We address these issues with VALA, a robust and efficient training-free framework that improves open-vocabulary grounding through visibility-aware gating (for contribution-aligned supervision) and robust multi-view aggregation.

\section{Preliminaries}
We briefly recall 3DGS~\cite{kerbl20233d} and how the features are assigned to a 3D Gaussian without iterative training.

% \noindent
\textbf{3D Gaussian Primitives and Projection.}
A scene is represented by a set of anisotropic Gaussians $\mathcal{G}=\{g_i\}_{i=1}^N$, with each Gaussian featured with $g_i=(\boldsymbol\mu_i,\mathbf\Sigma_i,\mathbf{c}_i,o_i)$, where $\boldsymbol\mu_i\!\in\!\mathbb{R}^3$ and $\mathbf\Sigma_i\!\in\!\mathbb{R}^{3\times 3}$ are the mean position and covariance matrix $\mathbf{c}_i$ encodes appearance (\eg, RGB or spherical harmonics coefficients), and $o_i\!\in\!(0,1]$ is a base opacity.

Images are rasterized by splatting the Gaussians from near to far along the camera ray through pixel $u$, followed by front-to-back $\alpha$-blending the Gaussian contributions, as:
% \vspace{-1em}
\begin{align}
\alpha_i(\mathbf{u})
&= 1-\exp\!\big(o_i\rho_i(\mathbf{u})\big),
\label{eq:alpha}
\\
T_i(\mathbf{u})
&= \prod\nolimits_{j<i}\!\big(1-\alpha_j(\mathbf{u})\big),
\hspace{-1em}
\label{eq:transmittance}
\\
\mathbf{C}(\mathbf{u})
&= \sum\nolimits_{i} 
%\underbrace{
\alpha_i(\mathbf{u})\,T_i(\mathbf{u})
%}_{w_i(\mathbf{u})}
\;\mathbf{c}_i(\mathbf{u}),
\label{eq:composite_color}
\end{align}
where $\rho_i(\mathbf{u})$ is the projected 2D Gaussian density in screen space, with projected 2D mean $\tilde{\boldsymbol\mu}_i$ and covariance $\tilde{\mathbf\Sigma}_i$, and
\begin{equation}
\rho_i(\mathbf{u})
= \exp\!\Big(-\tfrac{1}{2}(\mathbf{u}-\tilde{\boldsymbol\mu}_i)^\top
\tilde{\mathbf\Sigma}_i^{-1}(\mathbf{u}-\tilde{\boldsymbol\mu}_i)\Big).
\end{equation}

We denote the \emph{marginal contribution} of $g_i$ to pixel $\mathbf{u}$ as
\begin{equation}
w_i(\mathbf{u}) \;=\; \alpha_i(\mathbf{u})\,T_i(\mathbf{u}).
\label{eq:weight_def}
\end{equation}

% \noindent
\textbf{Language Features Assignment via Direct Aggregation.}
Recent works~\cite{drsplat25, cheng2024occam} proposed to directly assign 2D language features to 3D Gaussians via weighted feature aggregation. To obtain training-free 3D language feature embeddings, Kim~\etal~\cite{drsplat25} pool per-pixel weights $w_i(I, r)$, defined as in Eq.~\eqref{eq:weight_def}, using segmentation masks $M_j(I, r)$:
\begin{equation}
w_{ij} = \sum\nolimits_{I \in \mathcal{I}} \sum\nolimits_{r \in \Omega_I} M_j(I, r) \cdot w_i(I, r),
\end{equation}
where $w_{ij}$ associates Gaussian $i$-mask $j$, and $\Omega_I$ is the pixel domain of image $I$. The final CLIP embedding for each $i$ is a weighted average over the mask-level embeddings $f_j^{\mathrm{map}}$:
\begin{equation}
f_i = \sum\nolimits_{j=1}^{M} \frac{w_{ij}}{\sum_{k=1}^{M} w_{ik}} f_j^{\mathrm{map}}.
\label{eq:feature_aggregation_mask}
% f_i = \frac{\sum_{j=1}^{M} w_{ij} f_j^{\mathrm{map}}}{\sum_{j=1}^{M} w_{ij}}.
\end{equation}

Although this mask-based aggregation is a straightforward way to lift CLIP features into 3D, it has a memory footprint that scales quadratically with the scene complexity. To overcome this limitation, we adopt Occam's LGS~\cite{cheng2024occam}'s probabilistic per-view aggregation strategy as our baseline. \cite{cheng2024occam} avoids explicit mask representations and dense weight storage, maintaining semantic consistency across views. So, the 3D feature $f_i$ for Gaussian $i$ becomes:
\begin{equation}
    f_i = \frac{\sum_{s \in \mathcal{S}_i} w^s_i f^s_i}{\sum_{s \in \mathcal{S}_i} w^s_i},
    \label{feature aggregation_view}
\end{equation}
where $\mathcal{S}_i$ is the views set where Gaussian $i$ is visible, $w^s_i$ is the marginal contribution of $i$ at its center projection in view $s$, and $f^s_i$ is the 2D feature at the corresponding pixel.

\section{Method}
We aim to distill language features into 3DGS under \emph{visibility constraints}, to get semantically rich and \emph{view-consistent} 3D embeddings. 
Existing approaches that indiscriminately assign identical 2D features to \textit{all} Gaussians along a camera ray, which leads to noisy supervision and cross-view inconsistencies. With VALA, we assign only visible features.

Our pipeline is shown in \figureautorefname~\ref{fig:main_method}. Built on a direct feature assignment method, VALA has two complementary components to improve the assignment of 2D vision-language features to the 3D scene.
First, we introduce a \emph{visibility-aware attribution} mechanism to selectively assign language features to Gaussians based on their relevance in the rendered scene (Section~\ref{sec:gating}). 
Second, we propose a \emph{robust cross-view consolidation} strategy to aggregate per-view features while suppressing inconsistent observations, yielding coherent 3D semantic embeddings (Section~\ref{sec:median}).

\subsection{Visibility-Aware Feature Lifting}
\label{sec:gating}

\begin{figure}[tbp]
  \centering
  \vspace{-3mm}
  \includegraphics[width=0.95\linewidth]{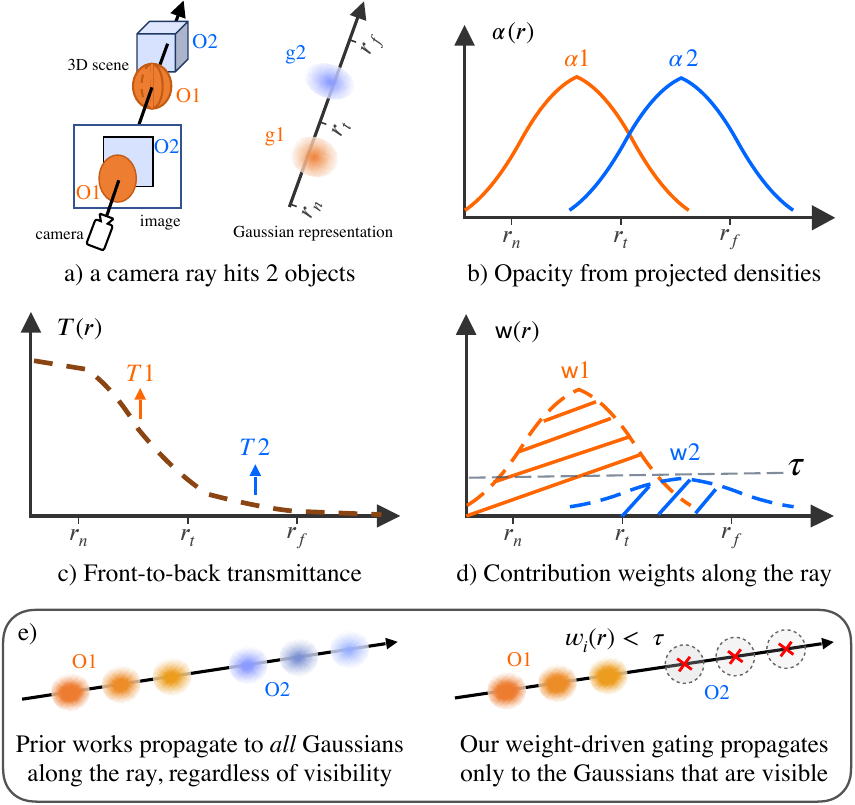} % png/jpg/pdf 都行
  \caption{\textbf{Visibility-aware gating for semantic assignment} (Section~\ref{sec:gating}). Simplified representation of a scene with two objects (a) \(O_1,O_2\) and a camera ray \(r\) with Gaussians \(g_1,g_2\). We compute the opacity (b) and compute the transmittance front-to-back (c). Then we calculate the contribution weights for each ray, thresholding with $\tau$ (d). Instead of propagating the features to \textit{all} Gaussians as prior works do, our gating only propagates to the visible ones (e).}
% (b) Per-Gaussian opacity \(\alpha_i(r)\) from projected densities.
% (c) Front-to-back transmittance \(T_i(r)\).
% (d) Per-ray contribution weights \(w_i(r)=\alpha_i(r)\,T_i(r)\).
% (e) We apply a threshold \(\tau\) and supervise only Gaussians with \(w_i(r)\ge \tau\), preventing the classic ray-wise semantic misassignment (assigning the same 2D feature to all Gaussians along a ray) and improving multi-view consistency.
\vspace{-5mm}
  \label{fig:weight_driven_gating}
\end{figure}

Recent works explored lifting 2D language embeddings into 3D space via differentiable rendering pipelines~\cite{cheng2024occam,drsplat25}. However, existing approaches assign the same 2D language feature to \textit{all} Gaussians intersected by a given pixel ray, regardless of each Gaussian’s actual contribution to the rendered pixel. As illustrated in Figure \ref{fig:weight_driven_gating}, when an object $O_2$ is occluded by another object $O_1$, the 2D language embedding at that pixel primarily represents the semantics of $O_1$. Nevertheless, a Gaussian $g_2$ belonging to $O_2$ may still be incorrectly associated with the language feature of $O_1$.

This erroneous assignment occurs in both alpha-blending-based language assignment methods~\cite{qin2024langsplat, liang2024supergseg} and, more prominently, in direct feature assignment methods~\cite{wu2024opengaussian, cheng2024occam, drsplat25}. As shown in Figure \ref{fig:weight_driven_gating} (b–c), even though the transmittance (Eq.~\eqref{eq:transmittance}) decreases monotonically along the ray from near to far—resulting in a very small transmittance for $g_2$—its alpha value (Eq.~\eqref{eq:alpha}) can remain relatively large in the far region. This, yields a non-negligible compositing weight (Eq.~\eqref{eq:visibility_weights}) for $g_2$, which, according to Eq.~\eqref{eq:feature_aggregation_mask} or Eq.~\eqref{feature aggregation_view}, contributes substantially to the final aggregated feature of $g_2$. Such unintended contributions introduce ambiguity into the 3D representation.

Recent works have introduced changes that indirectly affect this assignment.
Dr.Splat~\cite{drsplat25} selects the top-$k$ Gaussians along each ray, but this reduces computational costs rather than ensuring the correct semantic allocation. VoteSplat~\cite{jiang2025votesplat} recognizes that distant Gaussians may suffer from occlusion, but discards the compositing weights altogether and instead averages the features of \textit{all} intersected Gaussians to generate 3D votes for the clustering step. While they may tangentially bring improvements, they leave unsolved the assignment problem described above and continue to \textit{propagate wrong features} to background regions.

To overcome this limitation, we introduce a visibility-aware gating mechanism, which selectively supervises only the Gaussians along each ray that contribute to the pixel. By leveraging per-ray visibility weights, our method filters out occluded or low-contribution Gaussians before aggregating the features, ensuring that only geometrically and photometrically relevant points receive semantic supervision.
First, we clarify how we compute the \emph{per-ray weights}.

% \noindent
\textbf{Ray Notation and Marginal Contributions.}
Let $r$ denote the camera ray through pixel $\mathbf{u}$. For brevity, we write
\begin{equation}
\label{eq:visibility_weights}
% more space in 2 lines (because the ID (9) would go in the new line)
\begin{aligned}
T_i(r) &\equiv T_i(\mathbf{u}), 
\quad\quad\quad\quad \alpha_i(r) \equiv \alpha_i(\mathbf{u}), \\
w_i(r) &\equiv \alpha_i(r)\,T_i(r).
\end{aligned}
% packed version all in 1 line including the ID
% \begin{alignedat}{3}
% T_i(r) &\equiv T_i(\mathbf{u}), \hspace{0.35em}&
% \alpha_i(r) &\equiv \alpha_i(\mathbf{u}), \hspace{0.35em}&
% w_i(r) &\equiv \alpha_i(r)\,T_i(r).
% \end{alignedat}
\end{equation}
where $\alpha_i(r)$ encodes \emph{coverage} (\ie, how much $g_i$ overlaps the pixel), $T_i(r)$ represents \emph{transmittance} (\ie, how much light reaches $g_i$ after occlusion by nearer Gaussians), and $w_i(r)$ measures how strongly $g_i$ influences the rendered sample along $r$. We name this as the \emph{Visibility} of a Gaussian from a specific view. Instead of assigning this feature to all Gaussians on the ray $r$, we use a two-stage visibility-aware gate (VAG).
We aggregate the weights into a per-view visibility score
\begin{equation}
% S_i^{s} \;=\; \sum_{\mathbf r\in\Omega_s} w_i^{s}(\mathbf r),
% \qquad
S_{\mathrm{tot}}^s \;=\; \sum\nolimits_{i, r} w_i(r) .
\label{eq:view_score}
\end{equation}

% \noindent
\textbf{Stage A: Mass Coverage on the Thresholded Set.}  
We sort $\{w_i(r)\}_i$ decreasingly, with the indices as $(1),\ldots,(k)$. 
We then retain the shortest prefix that accounts for a target fraction $\tau_{\mathrm{view}}\!\in\![0.5,0.75]$ of the total visibility mass:
\begin{equation}
k_{\mathrm{mass}}^\star
\,=\,
\min\Big\{k:\ \sum\nolimits_{j=1}^{k} w_{j} \ \ge\ \tau_{\mathrm{view}}\; S_{\mathrm{tot}}^{s} \Big\}.
\end{equation}
To suppress numerical noise, we apply a small absolute floor $\tau_{\mathrm{abs}}$ and define the candidate set as
\begin{equation}
\mathcal{G}_{\mathrm{mass}}^{s}
=
\Big\{ (1),\ldots,(k_{\mathrm{mass}}^\star) \Big\}
\ \cap\
\big\{ i:\ w_i \ge \tau_{\mathrm{abs}} \big\}.
\label{eq:mass_floor}
\end{equation}

\textbf{Stage B: Quantile-Constrained Truncation.}
Let $\tau_{q}^s=\mathrm{Quantile}_{1-q}(\{w_i\}_i)$, we define 
$K_q^{s}=|\{ i:\ w_i \ge \tau_{q}^s \}|$ and instead of imposing a separate hard limit, we determine the selection cap directly via the $q$-quantile as
%. Specifically,  set the final prefix length
\begin{equation}
\begin{aligned}
k_{\mathrm{keep}}^{\star} & = \min\!\big(k_{\mathrm{mass}}^\star,\; K_q^{s}\big), \\
% \quad
\mathcal G_{\mathrm{keep}}^{s} & = \big\{(1),\ldots,(k_{\mathrm{keep}}^{\star})\big\}.
\label{eq:quantile_truncate}
\end{aligned}
\end{equation}

\textbf{Why Mass \emph{then} Quantile?}
A fixed quantile alone tightly controls cardinality but ignores how visibility mass is distributed,
and under heavy tails may discard essential contributors.
Conversely, mass coverage secures a target fraction of visible content but can be liberal when scores are flat.
Our two-stage rule reconciles both: Stage~A guarantees coverage on the \emph{relevant} (floored) set,
while B imposes a quantile-derived \emph{cardinality constraint} $K_q^{s}$ that stabilizes scale across views.
Practically, if $K_q^{s}\!\ge\!k_{\mathrm{mass}}^\star$, we keep the mass-coverage set unchanged;
otherwise we truncate it to the top-$K_q^{s}$ by $w_i$.
The gate is thus \emph{coverage-faithful} and \emph{scale-adaptive}.

\subsection{Robust Multi-View Aggregation}
\label{sec:median}

\begin{table*}[t]
  \centering
  \setlength{\tabcolsep}{6.8pt}
  \begin{tabular}{c l cc*{4}{cc}}
    \toprule
    & &
    \multicolumn{2}{c}{\textbf{Mean}} &
    \multicolumn{2}{c}{Figurines} &
    \multicolumn{2}{c}{Ramen} &
    \multicolumn{2}{c}{Teatime} &
    \multicolumn{2}{c}{Waldo\_Kitchen} \\
    \cmidrule(lr){3-4} \cmidrule(lr){5-6} \cmidrule(lr){7-8}
    \cmidrule(lr){9-10} \cmidrule(lr){11-12}
    & Method & mIoU & mAcc & mIoU & mAcc & mIoU & mAcc & mIoU & mAcc & mIoU & mAcc \\
    %\midrule[1.1pt]
    % \rowcolor{gray!12}\multicolumn{12}{l}{\textbf{2D evaluation}}\\[-2pt]
    \midrule
    \multirow{8}{*}{\rotatebox{90}{2D evaluation}}
    & LERF~\cite{kerr2023lerf}          & 37.4 & 73.6 & 38.6 & 75.0 & 28.2 & 62.0 & 45.0 & 84.8 & 37.9 & 72.7 \\
    & LEGaussian~\cite{shi2024language} & 24.6 & 67.4 & 23.4 & 57.1 & 20.2 & 69.0 & 32.3 & 79.7 & 22.3 & 63.6 \\
    & GOI~\cite{goi2024}                & 42.0 & 59.2 & 23.9 & 44.6 & 33.7 & 56.3 & 55.8 & 67.8 & 54.5 & 68.2 \\
    & GAGS~\cite{peng2024gags}          & 54.1 & 81.7 & 53.6 & 78.6 & 46.8 & 69.0 & 60.3 & 88.1 & 55.8 & 90.9 \\
    & LangSplat~\cite{qin2024langsplat} & 51.4 & \nd84.3 & 44.7 & 80.4 & 51.2 & 73.2 & 65.1 & 88.1 & 44.5 & \fr95.5 \\
    & LangSplatV2~\cite{li2025langsplatv2}   & 59.9 & 84.1 & 56.4 & \nd82.1 & \fr51.8 & \nd74.7 & \fr72.2 & \fr93.2 & 59.1 & 86.4 \\
    & Occam's LGS~\cite{cheng2024occam}                     & \nd61.3 & 82.5 & \nd58.6 & 80.4 & 51.0 & 74.7 & 70.2 & \nd93.2 & \fr65.3 & 81.8 \\
    & \textbf{VALA [ours]}                         & \fr61.7 & \fr86.4 & \fr59.9 & \fr82.1 & \nd51.5 & \fr75.6 & \nd70.2 & 91.5 & \nd65.1 & \nd86.4 \\
    %\midrule[1.1pt]
    % \rowcolor{gray!12}\multicolumn{12}{l}{\textbf{3D evaluation}}\\[-2pt]
    \midrule
    \multirow{10}{*}{\rotatebox{90}{3D evaluation}}
    & LangSplat~\cite{qin2024langsplat} & 10.35 & 13.64 & 7.27 & 10.71 & 10.05 & 9.86 & 14.38 & 20.34 & 9.71 & 9.09 \\
    & LEGaussians~\cite{shi2024language}& 16.21 & 23.82 & 17.99 & 23.21 & 15.79 & 26.76 & 19.27 & 27.12 & 11.78 & 18.18 \\
    & OpenGaussian~\cite{wu2024opengaussian}
                                        & 38.36 & 51.43 & 39.29 & 55.36 & 31.01 & 42.25 & 60.44 & 76.27 & 22.70 & 31.82 \\
    & SuperGSeg~\cite{liang2024supergseg}
                                        & 35.94 & 52.02 & 43.68 & 60.71 & 18.07 & 23.94 & 55.31 & 77.97 & 26.71 & 45.45 \\
    & Dr.Splat~\cite{drsplat25}
                                        & 43.29 & 64.30 & 54.42 & 80.36 & 24.33 & 35.21 & 57.35 & 77.97 & 37.05 & 63.64 \\
    & InstanceGaussian~\cite{li2024instancegaussian}
                                        & 43.87 & 61.09 & 54.87 & 73.21 & 25.03 & 38.03 & 54.13 & 69.49 & 41.47 & 63.64 \\
    % & OpenGaussian~\cite{wu2024opengaussian}
    %                                     & 44.26 & 60.24 & 54.35 & 75.01 & 29.23 & 50.85 & 62.26 & 79.66 & 31.20 & 45.45 \\
    % & Dr.Splat~\cite{drsplat25} \TODO{check}        & 43.56 & 68.11 & 51.82 & 75.00 & 24.91 & 45.07 & 56.27 & 79.66 & 41.25 & 72.72 \\
    & CAGS~\cite{sun2025cags}           & \nd50.79 & 69.62 & 60.85 & 82.14 & 36.29 & 46.48 & \nd68.40 & 86.44 & 37.62 & 63.64 \\
    & VoteSplat~\cite{jiang2025votesplat}& 50.10 & 67.38 & \fr68.62 & \nd85.71 & \nd39.24 & \nd61.97 & 66.71 & 88.14 & 25.84 & 33.68 \\ 
    & Occam's LGS~\cite{cheng2024occam}& 47.22 & \nd74.84 & 52.90 & 78.57 & 32.01 & 54.92 & 61.02 & \fr93.22 & \nd42.95 & \nd72.72 \\
    & \textbf{VALA [ours]}                      & \fr58.02 & \fr82.85 & \nd60.38 & \fr89.29 & \fr45.41 & \fr67.61 & \fr70.61 & \nd88.14 & \fr55.71 & \fr86.36 \\
    \bottomrule
  \end{tabular}
  \vspace{-2mm}
    \caption{Comparison on LERF-OVS (mIoU / mAcc.). 
    In 3D, results are taken from~\cite{wu2024opengaussian,liang2024supergseg,drsplat25,sun2025cags,jiang2025votesplat} and otherwise evaluated by us.
    %$\dagger$ means the results reported in the corresponding paper; * means our implementation.
    }
    \vspace{-5mm}
    \label{tab:lerfovs}
\end{table*}

SAM+CLIP preprocessing pipelines~\cite{qin2024langsplat} yield crisp mask boundaries and per-pixel open-vocabulary embeddings, but their semantics are often viewpoint-dependent: changes in viewpoint and occlusion induce noticeable drift across views. To enforce multi-view consistency, several 3DGS-based methods first form 3D-consistent clusters, typically supervised with contrastive signals derived from SAM masks, and then assign a language embedding to each cluster~\cite{wu2024opengaussian, liang2024supergseg, li2024instancegaussian, piekenbrinck2025opensplat3d}. While this decoupled clustering can improve multi-view semantic consistency, it makes the pipelines' training multi-stage, thus prolonging the training time. More critically, because clustering is still driven by noisy, view-dependent 2D cues, it does not correct the root cause, namely, upstream semantic drift, which can bias the clusters and ultimately degrade the accuracy of the final language assignments.

To address this multi-view inconsistency at source, we adopt geometric median~\cite{martini1995torricelli, weiszfeld2008english, beck2015weiszfeld} to robustly aggregate multi-view features by minimizing the cosine distances in feature space. Unlike aggregation by weighted mean, it dampens view-dependent outliers and semantic drift.

\textbf{Weighted Euclidean Geometric Median.}
Using the visibility weights defined in Eq.~\eqref{eq:visibility_weights}, the (weighted) geometric median for $g_i$ is
\begin{equation}
\label{eq:gm_euclid}
\mathbf{z}_i^\star
=
\text{argmin}_{\mathbf{z}\in\mathbb{R}^d}
\;\sum\nolimits_{s} 
w_i(r)\,\bigl\lVert \mathbf{z} - \mathbf{f}_i^s \bigr\rVert .
\end{equation}

\textbf{Cosine-loss Median on the Unit Sphere.}
$\mathbf{f}(I,\mathbf{u})$ are $\ell_2$-normalized embeddings and thus angular consistency is most relevant. Therefore, we constrain $\mathbf{z}_i$ to the unit sphere $\mathbb{S}^{d-1}$ and minimize a weighted cosine loss:
\vspace{-3mm}
\begin{equation}
\label{eq:cosine_obj}
\mathbf{z}_i^\star 
\,=\, \text{argmin}_{\lVert \mathbf{z}\rVert_2=1}
\;\sum\nolimits_{s}
w_i(r)\,\bigl( 1 - \mathbf{f}_i^{s\top} \mathbf{z} \bigr),
\end{equation}
where $w_i(r)$ denotes the visibility weight of Gaussian $g_i$ from view $s$, since $r$ represents the view $s$.
The gradient of $\ell(\mathbf{f},\mathbf{z})=1-\mathbf{f}^\top\mathbf{z}$ on $\mathbb{S}^{d-1}$ is
\(
\nabla_{\mathbf{z}} \ell = -\bigl[\mathbf{f}-(\mathbf{f}^\top\mathbf{z})\,\mathbf{z}\bigr],
\)
the projection of $\mathbf{f}$ onto the tangent space at $\mathbf{z}$.
Compared to the Euclidean formulation in Eq.~\eqref{eq:gm_euclid}, this objective directly optimizes angular similarity, circumventing the scale sensitivity of Euclidean distances in high dimensions, where norm variations dominate over angular differences, and empirically leads to more stable 3D semantics (Table~\ref{tab:ablation_lerf}). 
% which avoids scale issues in high dimensions and empirically yields more stable 3D semantics.

% \vspace{-3mm}
\begin{algorithm}[t!]
\caption{Streaming cosine-loss median on $\mathbb{S}^{d-1}$ (Section~\ref{sec:median}).}
% \vspace{-3mm}
\label{alg:streaming_cosine_median}
\begin{algorithmic}[1]
\Require Stream $\{(\mathbf{f}_t, w_i^t)\}_{t=1}^T$ with $\mathbf{f}_t\in\mathbb{R}^d$, $\|\mathbf{f}_t\|_2=1$, and $w_i^t>0$
\State Initialize $\mathbf{z}_{i,0}\leftarrow \mathbf{f}_1$, \quad $W_{i,0}\leftarrow 0$
\For{$t=1,\ldots,T$}
    \State $\mathbf{d}_t \leftarrow \mathbf{f}_t - (\mathbf{f}_t^\top \mathbf{z}_{i,t})\,\mathbf{z}_{i,t}$ \Comment{tangent direction}
    \State $\eta_t \leftarrow \dfrac{w_i^t}{W_{i,t}+w_i^t}$ \Comment{streaming step size}
    \State $\mathbf{z}_{i,t+1} \leftarrow \mathrm{Norm}\!\big(\mathbf{z}_{i,t} + \eta_t\,\mathbf{d}_t\big)$
    \State $W_{i,t+1} \leftarrow W_{i,t} + w_i^t$
\EndFor
\State \Return $\mathbf{z}_i \leftarrow \mathbf{z}_{i,T}$, \ $W_i \leftarrow W_{i,T}$
% \vspace{-5mm}
\end{algorithmic}
% \vspace{-5mm}
\end{algorithm}
% \vspace{-1mm}

% \noindent
\textbf{Constant-Memory Streaming Update.} 
While effective, solving Eq.~\eqref{eq:cosine_obj} with the classical Weiszfeld algorithm~\cite{eckhardt1980weber} requires repeated full-batch updates over all Gaussian features, which scales linearly with the number of views and becomes computationally prohibitive in practice. 
To address this, we adopt a constant-memory streaming scheme inspired by online optimization~\cite{kirillov2023sam}.
Specifically, as detailed in Algorithm~\ref{alg:streaming_cosine_median}, we maintain only the current estimate $(\mathbf{z}_{i,t},W_{i,t})$, where $W_{i,t}$ is the cumulative visibility weight, and incorporate each new observation $(\mathbf{f}_t,w_i^t)$ via
\vspace{-1mm}
\begin{align}
\mathbf{z}_{i,t+1}
&= \mathrm{Norm}\!\left(
\mathbf{z}_{i,t}
+\eta_t\, w_i^t\,
\bigl[\mathbf{f}_t - (\mathbf{f}_t^\top \mathbf{z}_{i,t})\,\mathbf{z}_{i,t}\bigr]
\right), \\
\eta_t &= \frac{w_i^t}{W_{i,t}+w_i^t}\,, \qquad
W_{i,t+1} = W_{i,t} + w_i^t,
\vspace{-2mm}
\end{align}

% where $\mathrm{Norm}(\mathbf{x})=\mathbf{x}/\|\mathbf{x}\|_2$ ensures $\mathbf{z}_{i,t}\in\mathbb{S}^{d-1}$ at every step.
% The direction $\mathbf{f}_t-(\mathbf{f}_t^\top \mathbf{z}_{i,t})\mathbf{z}_{i,t}$ corresponds to the tangent component that increases cosine similarity, while the adaptive step size $\eta_t$ ensures each sample contributes proportionally to its visibility.
% Under standard stochastic approximation assumptions (bounded variance and diminishing step sizes), $\mathbf{z}_{i,t}$ converges to a stationary point of Eq.~\eqref{eq:cosine_obj} at rate $\mathcal{O}(1/\sqrt{W_{i,t}})$.
where $\mathrm{Norm}(\mathbf{x})=\mathbf{x}/\|\mathbf{x}\|_2$ projects $\mathbf{z}_{i,t}$ onto the unit sphere $\mathbb{S}^{d-1}$.
The update direction $\mathbf{f}_t-(\mathbf{f}_t^\top \mathbf{z}_{i,t})\mathbf{z}_{i,t}$ lies in the tangent space and increases cosine similarity, while the adaptive step size $\eta_t$ weights each sample according to its visibility.
Under standard stochastic approximation assumptions, $\mathbf{z}_{i,t}$ converges to a stationary point of Eq.~\eqref{eq:cosine_obj} at rate $\mathcal{O}(1/\sqrt{W_{i,t}})$.

\section{Experiments}

\label{sec:exp}
% In this section, we first introduce datasets and our novel evaluation protocol on ScanNet, which preserves not only the rendering quality but also provides a fair comparison on 3D point clouds via a novel semantic propagation method.
\subsection{Experimental setup}
\label{exp:setup}

\textbf{Datasets.}
We evaluate on the two reference datasets for this task: LERF-OVS~\cite{qin2024langsplat} and ScanNet-v2~\cite{dai2017scannet}. LERF-OVS is derived from the LERF dataset of Kerr \etal~\cite{kerr2023lerf}, where we evaluate open-vocabulary object selection in both 2D and 3D. 
For the 2D evaluation, we follow the protocol of LERF~\cite{kerr2023lerf}. For the 3D evaluation, we follow OpenGaussian~\cite{wu2024opengaussian}. On ScanNet, we evaluate 3D semantic segmentation. Previous evaluation protocols~\cite{wu2024opengaussian,li2024instancegaussian} freeze the growth of 3D Gaussians, which degrades photometric fidelity. In contrast, we allow full optimization of the 3D Gaussians, resulting in misalignment between the optimized Gaussians and the ground-truth point cloud. We therefore adapt the evaluation protocol in~\cite{drsplat25} by propagating pseudo ground-truth labels to the Gaussians. Details are provided in the Appendix~\ref{subsec:eval_prot}. 
% All results in each table follow the same protocol.

\textbf{Implementation Details.} 
We generate SAM~\cite{kirillov2023sam} masks at subpart, part, and whole object granularities. We use OpenCLIP ViT-B/16~\cite{radford2021clip} and the \texttt{gsplat} rasterizer~\cite{ye2025gsplat}. We apply direct feature aggregation in the 512-dimensional space following~\cite{cheng2024occam}, combined with our proposed training-free method. The entire process requires only 10 seconds to one minute per scene (depending on scene scale), thanks to our effective cross-view feature aggregation and streaming updates at constant memory. For all experiments, we used an NVIDIA RTX 4090 GPU.

\begin{figure*}[t]
\centering
\includegraphics[width=1.\linewidth]{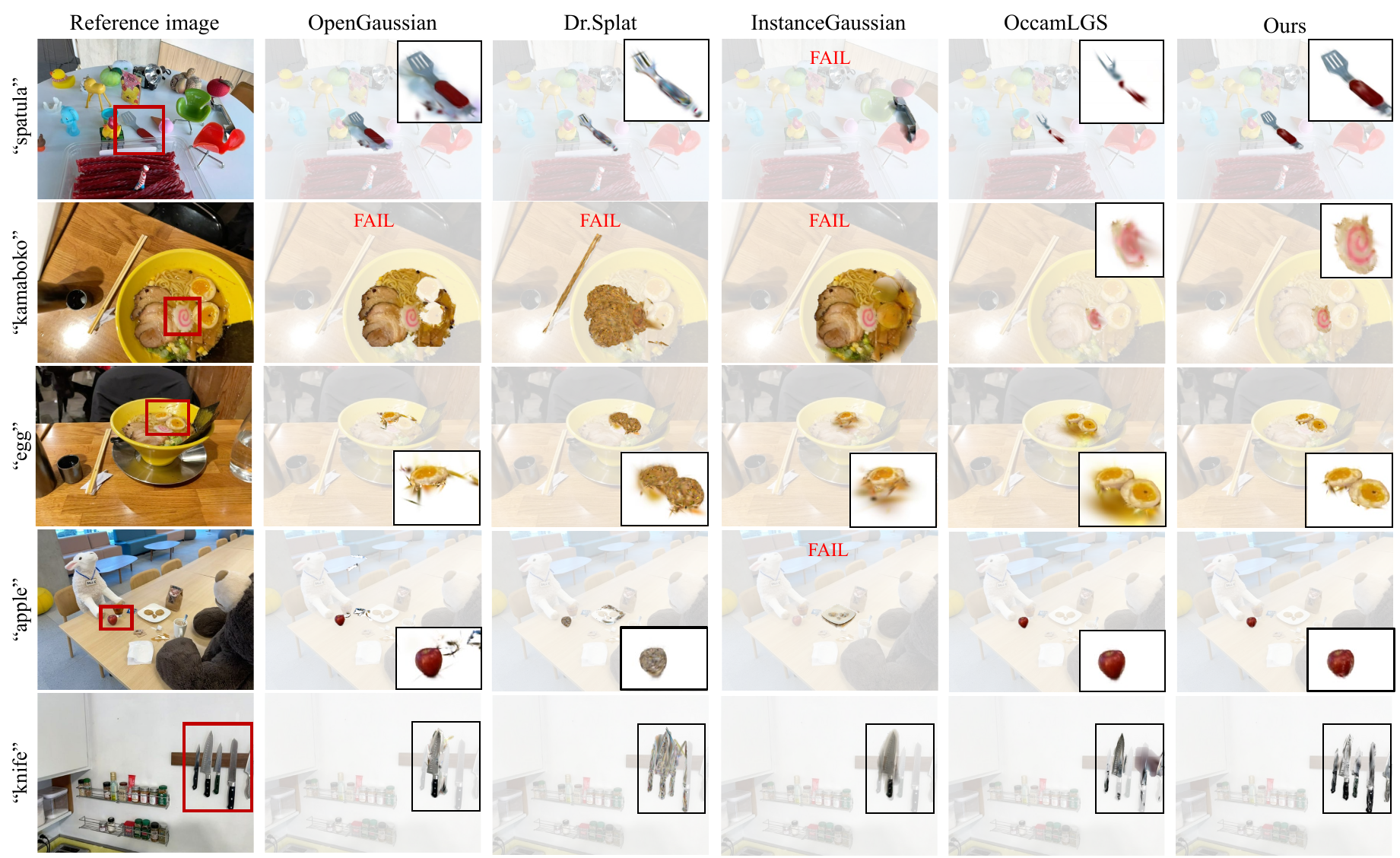}
\vspace{-5mm}
\caption{Qualitative 3D objects selections on LeRF-OVS~\cite{qin2024langsplat}. We mark as failed those with low or zero IoU with the ground truth (red).}
\label{fig:qualitative_lerfovs}
\vspace{-3mm}
\end{figure*}

\subsection{Analysis on LeRF-OVS dataset}
\tableautorefname~\ref{tab:lerfovs} compares ours with state-of-the-art works on LERF-OVS in 2D and 3D. In 2D, per-view segmentation quality projected from 3D is checked, while in 3D, we directly assess multi-view consistent semantic reconstruction.

\textbf{Quantitatives In 2D.} Our method achieves the highest scores on both mIoU and mAcc, slightly surpassing the mIoU of Occam's LGS~\cite{cheng2024occam} and outperforming LangSplatV2~\cite{li2025langsplatv2}. This improvement is consistent across diverse scenes, particularly in Figurines and Ramen, suggesting that our visibility-aware attribution reduces per-ray semantic noise without sacrificing fine-grained per-view accuracy. While GAGS~\cite{peng2024gags} and LangSplat~\cite{qin2024langsplat} also deliver competitive 2D scores, their performance drops with complex occlusions (\eg, Ramen for GAGS), indicating that their 2D-driven assignments do not fully mitigate cross-view inconsistencies.

\textbf{Quantitatives In 3D.} The advantage of our method becomes more pronounced in 3D, with ours exceeding all baselines by a notable margin. The second best, CAGS~\cite{sun2025cags}, is a substantial 7.2 absolute mIoU points behind. The scene-level analysis reveals that our approach leads in Ramen, Teatime, and Waldo\_Kitchen, and ranks second in Figurines, behind VoteSplat~\cite{jiang2025votesplat} due to its specialized multi-view voting. The gains are especially significant in large, cluttered environments (Teatime, Waldo\_Kitchen), where our contribution-aware aggregation better preserves semantics despite severe occlusions.

The strong 3D consistency of our method contrasts with approaches like LangSplat and LEGaussian~\cite{shi2024language}, whose high 2D accuracy does not translate to 3D performance, likely due to their lack of explicit handling of per-ray contribution and occlusion. Similarly, the post-hoc clustering methods OpenGaussian~\cite{wu2024opengaussian} and SuperGSeg~\cite{liang2024supergseg} exhibit moderate 3D improvements but remain sensitive to upstream semantic drift, thereby limiting their robustness. Our performance relative to Occam's LGS (baseline) is noteworthy: while both adopt streaming updates, our visibility-guided feature attribution yields much better performance in 3D, highlighting the effectiveness of improving the semantic assignment at the feature aggregation stage rather than solely relying on memory-efficient training.

\textbf{Qualitatives in 3D.} 
We show visual 3D results in Figure~\ref{fig:qualitative_lerfovs}. %We evaluate the 3D object selection task on the LeRF-OVS dataset. 
Existing approaches, such as InstanceGaussian~\cite{li2024instancegaussian}, frequently fail by retrieving incorrect objects across multiple scenes. This can be attributed to their reliance on appearance–semantic joint representations, which struggle to distinguish small objects with visually similar appearances. Clustering-based methods struggle with multiple instances that are closely related. For example, querying for \emph{``knife''}, OpenGaussian~\cite{wu2024opengaussian} and InstanceGaussian~\cite{li2024instancegaussian} detect only one out of five knives, whereas Dr.Splat~\cite{drsplat25} and Occam’s LGS~\cite{cheng2024occam} identify all knives but produce indistinct boundaries. In contrast, ours successfully localizes all knives with accurate and sharp delineations. Our approach also demonstrates robustness on challenging small-object queries, such as \emph{``Kamaboko''} and \emph{``egg''} in the \emph{Ramen} scene. These targets lie within a heavily cluttered context (a bowl of ramen), making them particularly difficult to isolate. Competing methods~\cite{wu2024opengaussian, drsplat25, li2024instancegaussian} fail to recognize these objects, while Occam’s LGS correctly retrieves them but with blurred contours. By comparison, ours produces precise boundaries and accurately captures fine object structures. Similar improvements are observed in the \emph{``Spatula''} query, further illustrating that our visibility-aware gating not only mitigates occlusion effects but also enables the recovery of fine-grained details in complex scenes.

\subsection{3D Semantic Segmentation on ScanNet}

\begin{figure*}[t]
\centering
\includegraphics[width=\linewidth]{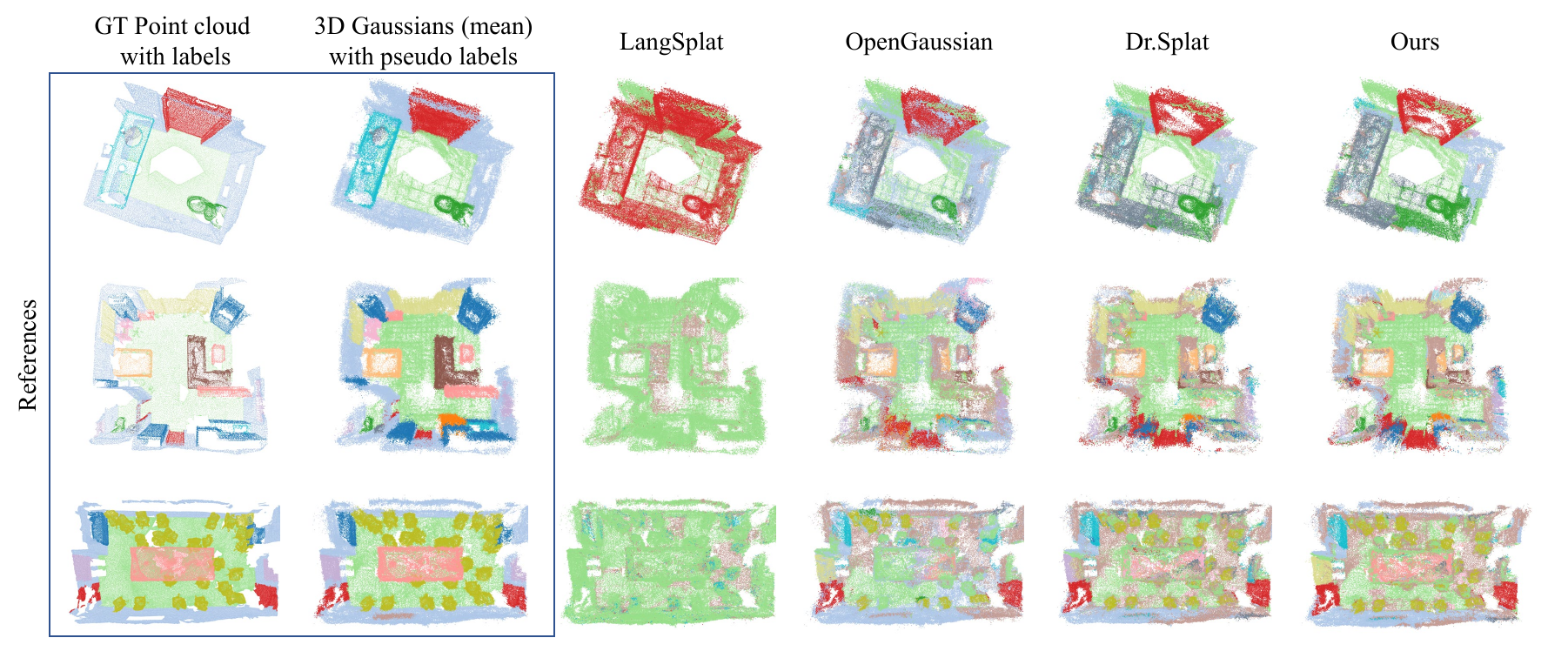}
\vspace{-9mm}
\caption{Qualitative results of 3D semantic segmentation with 19 classes on the ScanNet-v2 dataset~\cite{dai2017scannet}.}
\label{fig:qualitative_scannet}
~\vspace{-9mm}
\end{figure*}

\begin{table}[t]
\centering
\small
\setlength{\tabcolsep}{2.85pt}
\renewcommand{\arraystretch}{1.1}
\begin{tabular}{lcccccc}
\toprule
& \multicolumn{2}{c}{19 classes} & \multicolumn{2}{c}{15 classes} & \multicolumn{2}{c}{10 classes} \\
\cmidrule(lr){2-3} \cmidrule(lr){4-5} \cmidrule(lr){6-7}
Method & mIoU & mAcc & mIoU & mAcc & mIoU & mAcc \\
\midrule
LangSplat~\cite{kerr2023lerf} & 2.45 & 8.59 & 3.45 & 13.21 & 6.48 & 21.89 \\
OpenGaussian~\cite{wu2024opengaussian} & 27.73 & 42.01 & 29.67 & 46.15 & 39.93 & 57.34 \\
Dr. Splat~\cite{drsplat25} & 29.31 & 47.68 & 33.25 & \nd54.33 & 44.19 & \nd65.19 \\
Occam's LGS~\cite{cheng2024occam} & \nd31.93 & \nd48.93 & \nd34.25 & 53.71 & \nd45.16 & 64.39 \\
\textbf{VALA [ours]} & \fr32.11 & \fr50.05 & \fr35.10 & \fr54.77 & \fr46.21& \fr65.61 \\
\bottomrule
\end{tabular}
\vspace{-2mm}
\caption{Open-vocabulary 3D semantic segmentation task on the ScanNet-v2 dataset~\cite{dai2017scannet} across different amounts of classes.}
\label{tab:openvocab_seg}
\vspace{-5mm}
\end{table}

\textbf{Quantitatives.}  
As reported in Table~\ref{tab:openvocab_seg}, our method achieves the best performance across all evaluation settings, including the most challenging 19-class scenario. Compared to Occam's LGS~\cite{cheng2024occam}, our contribution-aware aggregation is advantageous, demonstrating its ability to handle fine-grained class distributions. While Dr.Splat~\cite{drsplat25} attains competitive accuracy in reduced-category settings, it lags notably in mIoU, indicating weaker spatial consistency. These results confirm that our method achieves robust and precise 3D segmentation across varying label granularities.

% \noindent
\textbf{Qualitatives.}  
Qualitative comparisons are presented in \figureautorefname~\ref{fig:qualitative_scannet}. In the large and complex second room, our method accurately predicts the wall behind the bed (bed in orange), a structure often misclassified by others. In the smaller but more occluded third scene, our method also demonstrates superior 3D segmentation, capturing challenging objects such as the central table more effectively. This ability to recover occluded and fine-scale geometry is particularly beneficial for downstream applications such as 3D object localization. Overall, the qualitative results support the quantitative improvements, highlighting both the robustness and effectiveness of our proposed framework.

\vspace{-1mm}
\begin{table}[b]
  \centering
  \small
  \vspace{-5mm}
  \begin{tabular}{lccccc}
    \toprule
    Ref. & Stage A & Stage B & Median & mIoU & mAcc \\
    \cmidrule(lr){1-1} \cmidrule(lr){2-4} \cmidrule(lr){5-6}
    %Occam's LGS~\cite{cheng2024occam} (no gating + weighted mean)  
    O.LGS~\cite{cheng2024occam} & && & 47.22 & 74.86 \\
    % \midrule
    (b) & && cosine & 49.03 & 80.08 \\
    % Occam's LGS + our cosine median  & 49.03 & 80.08 \\
    (c) & \checkmark & & cosine              & 57.24 & 81.25 \\
    (d) &    & \checkmark & cosine  & 55.21 & 80.37 \\
    % first stage B then stage A gating & -- & -- \\
    % Stage A and Stage B gating &57.95&82.55\\
    % \textbf{Full: Stage A + Stage B (ours)}           & -- & -- \\
    \textbf{VALA} & \checkmark & \checkmark & cosine & \textbf{58.02}& \textbf{82.85}\\
    (f) &    \checkmark & \checkmark &  & 52.29 &76.17 \\
    (g) & \checkmark & \checkmark & L1 & 56.03 & 82.42 \\
    % gating + cosine median &  &  \\
    % no gating, with l1 distance    &45.83& 74.75\\
    \bottomrule
  \end{tabular}
  \vspace{-2mm}
  \caption{Ablation on LeRF-OVS. First row is Occam's LGS~\cite{cheng2024occam}, \ie, our baseline. Stages from Section~\ref{sec:gating}, Median from~\ref{sec:median}. All rows share the same data, rasterizer, and hyperparameters.}
  \label{tab:ablation_lerf}
  % \vspace{-5mm}
\end{table}
    
\subsection{Ablation Study}
% \lipsum[1]
We conduct an ablation study on LeRF-OVS~\cite{qin2024langsplat}, averaging the metrics over all scenes. Table~\ref{tab:ablation_lerf} disentangles the contributions of our main components, namely visibility-aware gating and cosine-based geometric median. Starting from the baseline Occam's LGS~\cite{cheng2024occam}, replacing the naive weighted mean with our cosine median (b) already improves performance, highlighting the advantage of robust aggregation in the embedding space. Incorporating visibility-aware gating further boosts results (c-d), where mass-coverage plus threshold gating (c) yields the strongest individual gain, while quantile pruning (d) provides complementary benefits. We also observe that our gating alone (f) is less effective compared to gating along with our robust median (VALA), showing that the precise aggregation is critical to fully exploit visibility cues. Lastly, we compare cosine and L1 (g) as median, with the former delivering superior results. Our full model (VALA) achieves the best overall performance, validating that both visibility-aware gating and cosine-based median aggregation are important for an accurate and view-consistent 2D-3D language lifting.

We refer to the \textbf{Supplementary Material} for additional details and results.
\section{Conclusion}
We introduced VALA, an efficient and effective method to address two fundamental problems in the feature aggregation of open-vocabulary recognition in 3DGS, namely (i) the propagation of 2D features to all Gaussians along a camera ray, and (ii) the multi-view inconsistency of semantic features. VALA tackles (i) with a visibility-aware distillation of language features based on a two-stage gating mechanism, and (ii) with a cosine variant of the geometric median, updating the features via streaming to keep the memory footprint low.
These innovations ensure more appropriate features are assigned to the 3D Gaussians, ultimately leading to superior performance in open-vocabulary segmentation. Remarkably, the proposed VALA achieves state-of-the-art performance on 2D \textit{and} 3D tasks on the reference datasets LeRF-OVS and ScanNet-v2.

{
    \small
    \bibliographystyle{ieeenat_fullname}
    \bibliography{main}

@String(CVPR= {IEEE Conf. Comput. Vis. Pattern Recog.})

@String(CVPR  = {CVPR})

@article{kerbl20233d,
  title={{3D Gaussian Splatting} for Real-Time Radiance Field Rendering.},
  author={Kerbl, Bernhard and Kopanas, Georgios and Leimk{\"u}hler, Thomas and Drettakis, George},
  journal={ACM Transactions on Graphics},
  volume={42},
  number={4},
  pages={139--1},
  year={2023}
}

@inproceedings{engelmann2024opennerf,
  title={{OpenNeRF}: Open Set {3D} Neural Scene Segmentation with Pixel-Wise Features and Rendered Novel Views},
  author={Engelmann, Francis and Manhardt, Fabian and Niemeyer, Michael and Tateno, Keisuke and Tombari, Federico},
  booktitle={The Twelfth International Conference on Learning Representations},
  year={2024}
}

@inproceedings{kerr2023lerf,
  title={Lerf: Language embedded radiance fields},
  author={Kerr, Justin and Kim, Chung Min and Goldberg, Ken and Kanazawa, Angjoo and Tancik, Matthew},
  booktitle={Proceedings of the IEEE/CVF International Conference on Computer Vision},
  pages={19729--19739},
  year={2023}
}

@inproceedings{radford2021clip,
  title={Learning transferable visual models from natural language supervision},
  author={Radford, Alec and Kim, Jong Wook and Hallacy, Chris and Ramesh, Aditya and Goh, Gabriel and Agarwal, Sandhini and Sastry, Girish and Askell, Amanda and Mishkin, Pamela and Clark, Jack and others},
  booktitle={Proceedings of the International Conference on Machine Learning},
  pages={8748--8763},
  year={2021},
  organization={PMLR}
}

@inproceedings{kirillov2023sam,
  title={Segment anything},
  author={Kirillov, Alexander and Mintun, Eric and Ravi, Nikhila and Mao, Hanzi and Rolland, Chloe and Gustafson, Laura and Xiao, Tete and Whitehead, Spencer and Berg, Alexander C and Lo, Wan-Yen and others},
  booktitle={Proceedings of the IEEE/CVF International Conference on Computer Vision},
  pages={4015--4026},
  year={2023}
}

@inproceedings{qin2024langsplat,
  title={Langsplat: {3D} language gaussian splatting},
  author={Qin, Minghan and Li, Wanhua and Zhou, Jiawei and Wang, Haoqian and Pfister, Hanspeter},
  booktitle={Proceedings of the IEEE/CVF Conference on Computer Vision and Pattern Recognition},
  pages={20051--20060},
  year={2024}
}

@article{li2025langsplatv2,
  title={LangSplatV2: High-dimensional 3D Language Gaussian Splatting with 450+ FPS},
  author={Li, Wanhua and Zhao, Yujie and Qin, Minghan and Liu, Yang and Cai, Yuanhao and Gan, Chuang and Pfister, Hanspeter},
  journal={arXiv preprint arXiv:2507.07136},
  year={2025}
}

@inproceedings{mildenhall2020nerf,
  title={{NeRF}: Representing Scenes as Neural Radiance Fields for View Synthesis},
  author={Mildenhall, Ben and Srinivasan, Pratul P and Tancik, Matthew and Barron, Jonathan T and Ramamoorthi, Ravi and Ng, Ren},
  booktitle={Proceedings of the European Conference on Computer Vision},
  pages={405--421},
  year={2020}
}

@article{liang2024supergseg,
  title={Supergseg: Open-vocabulary 3d segmentation with structured super-gaussians},
  author={Liang, Siyun and Wang, Sen and Li, Kunyi and Niemeyer, Michael and Gasperini, Stefano and Navab, Nassir and Tombari, Federico},
  journal={arXiv preprint arXiv:2412.10231},
  year={2024}
}

@inproceedings{zhou2024feature,
  title={Feature {3DGS}: Supercharging {3D} gaussian splatting to enable distilled feature fields},
  author={Zhou, Shijie and Chang, Haoran and Jiang, Sicheng and Fan, Zhiwen and Zhu, Zehao and Xu, Dejia and Chari, Pradyumna and You, Suya and Wang, Zhangyang and Kadambi, Achuta},
  booktitle={Proceedings of the IEEE/CVF Conference on Computer Vision and Pattern Recognition},
  pages={21676--21685},
  year={2024}
}

@article{peng2024gags,
  title={Gags: Granularity-aware feature distillation for language gaussian splatting},
  author={Peng, Yuning and Wang, Haiping and Liu, Yuan and Wen, Chenglu and Dong, Zhen and Yang, Bisheng},
  journal={arXiv preprint arXiv:2412.13654},
  year={2024}
}

@article{wu2024opengaussian,
  title={{OpenGaussian}: Towards point-level {3D} gaussian-based open vocabulary understanding},
  author={Wu, Yanmin and Meng, Jiarui and Li, Haijie and Wu, Chenming and Shi, Yahao and Cheng, Xinhua and Zhao, Chen and Feng, Haocheng and Ding, Errui and Wang, Jingdong and others},
  journal={Advances in Neural Information Processing Systems},
  volume={37},
  pages={19114--19138},
  year={2024}
}

@inproceedings{dai2017scannet,
    title={ScanNet: Richly-annotated {3D} Reconstructions of Indoor Scenes},
    author={Dai, Angela and Chang, Angel X. and Savva, Manolis and Halber, Maciej and Funkhouser, Thomas and Nie{\ss}ner, Matthias},
    booktitle = {Proceedings of the IEEE/CVF Conference on Computer Vision and Pattern Recognition},
    year = {2017}
}

@inproceedings{shi2024language,
  title={Language embedded {3D} gaussians for open-vocabulary scene understanding},
  author={Shi, Jin-Chuan and Wang, Miao and Duan, Hao-Bin and Guan, Shao-Hua},
  booktitle={Proceedings of the IEEE/CVF Conference on Computer Vision and Pattern Recognition},
  pages={5333--5343},
  year={2024}
}

@article{zuo2024fmgs,
  title={Fmgs: Foundation model embedded {3D} gaussian splatting for holistic {3D} scene understanding},
  author={Zuo, Xingxing and Samangouei, Pouya and Zhou, Yunwen and Di, Yan and Li, Mingyang},
  journal={International Journal of Computer Vision},
  pages={1--17},
  year={2024},
  publisher={Springer}
}

@inproceedings{qi2017pointnet,
  title={{PointNet}: Deep learning on point sets for {3D} classification and segmentation},
  author={Qi, Charles R and Su, Hao and Mo, Kaichun and Guibas, Leonidas J},
  booktitle={Proceedings of the IEEE/CVF Conference on Computer Vision and Pattern Recognition},
  pages={652--660},
  year={2017}
}

@article{sun2025cags,
  title={Cags: Open-vocabulary 3d scene understanding with context-aware gaussian splatting},
  author={Sun, Wei and Zhou, Yanzhao and Jiao, Jianbin and Li, Yuan},
  journal={arXiv preprint arXiv:2504.11893},
  year={2025}
}

@inproceedings{li2024instancegaussian,
  title={{InstanceGaussian}: Appearance-semantic joint gaussian representation for {3D} instance-level perception},
  author={Li, Haijie and Wu, Yanmin and Meng, Jiarui and Gao, Qiankun and Zhang, Zhiyao and Wang, Ronggang and Zhang, Jian},
  booktitle={Proceedings of the IEEE/CVF Conference on Computer Vision and Pattern Recognition},
  pages={14078--14088},
  year={2025}
}

@article{cheng2024occam,
  title={{Occam's LGS}: A Simple Approach for Language {Gaussian} Splatting},
  author={Cheng, Jiahuan and Zaech, Jan-Nico and Van Gool, Luc and Paudel, Danda Pani},
  journal={arXiv preprint arXiv:2412.01807},
  year={2024}
}

@inproceedings{drsplat25,
    title={{Dr. Splat}: Directly Referring {3D Gaussian Splatting} via Direct Language Embedding Registration},
    author={Jun-Seong, Kim and Kim, GeonU and Yu-Ji, Kim and Wang, Yu-Chiang Frank and Choe, Jaesung and Oh, Tae-Hyun},
    booktitle={Proceedings of the IEEE/CVF Conference on Computer Vision and Pattern Recognition},
    year={2025}
}

@inproceedings{goi2024,
  title={{GOI}: Find {3D} gaussians of interest with an optimizable open-vocabulary semantic-space hyperplane},
  author={Qu, Yansong and Dai, Shaohui and Li, Xinyang and Lin, Jianghang and Cao, Liujuan and Zhang, Shengchuan and Ji, Rongrong},
  booktitle={Proceedings of the ACM International Conference on Multimedia},
  pages={5328--5337},
  year={2024}
}

@article{jiang2025votesplat,
  title={VoteSplat: Hough Voting Gaussian Splatting for 3D Scene Understanding},
  author={Jiang, Minchao and Jia, Shunyu and Gu, Jiaming and Lu, Xiaoyuan and Zhu, Guangming and Dong, Anqi and Zhang, Liang},
  journal={arXiv preprint arXiv:2506.22799},
  year={2025}
}

@inproceedings{piekenbrinck2025opensplat3d,
  title={OpenSplat3D: Open-Vocabulary 3D Instance Segmentation using Gaussian Splatting},
  author={Piekenbrinck, Jens and Schmidt, Christian and Hermans, Alexander and Vaskevicius, Narunas and Linder, Timm and Leibe, Bastian},
  booktitle={Proceedings of the Computer Vision and Pattern Recognition Conference},
  pages={5246--5255},
  year={2025}
}

@article{zhan2025hi,
  title={Hi-LSplat: Hierarchical 3D Language Gaussian Splatting},
  author={Zhan, Chenlu and Zhang, Yufei and Wang, Gaoang and Wang, Hongwei},
  journal={arXiv preprint arXiv:2506.06822},
  year={2025}
}

@article{martini1995torricelli,
  author  = {Martini, Horst and Swanepoel, Konrad J. and Weiss, G{\"u}nter},
  title   = {On Torricelli's geometrical solution to a problem of Fermat},
  journal = {Elemente der Mathematik},
  year    = {1995},
  volume  = {50},
  number  = {2},
  pages   = {93--96}
}

@article{weiszfeld2008english,
  author    = {Weiszfeld, Endre and Plastria, Frank},
  title     = {On the point for which the sum of the distances to $n$ given points is minimum},
  journal   = {Annals of Operations Research},
  year      = {2008},
  volume    = {167},
  number    = {1},
  pages     = {7--41},
  doi       = {10.1007/s10479-008-0352-z}
}

@article{beck2015weiszfeld,
  author  = {Beck, Amir and Sabach, Shoham},
  title   = {Weiszfeld’s Method: Old and New Results},
  journal = {Optimization Letters},
  year    = {2015},
  volume  = {9},
  number  = {1},
  pages   = {1--18},
  note    = {See also preprint/PDF for historical notes}
}

@article{ye2025gsplat,
  title={gsplat: An open-source library for Gaussian splatting},
  author={Ye, Vickie and Li, Ruilong and Kerr, Justin and Turkulainen, Matias and Yi, Brent and Pan, Zhuoyang and Seiskari, Otto and Ye, Jianbo and Hu, Jeffrey and Tancik, Matthew and Angjoo Kanazawa},
  journal={Journal of Machine Learning Research},
  volume={26},
  number={34},
  pages={1--17},
  year={2025}
}

@article{cadena2016slam,
  title={Past, Present, and Future of Simultaneous Localization and Mapping: Toward the Robust-Perception Age},
  author={Cadena, Cesar and Carlone, Luca and Carrillo, Henry and Latif, Yasir and Scaramuzza, Davide and Neira, Jos{\'e} and Reid, Ian and Leonard, John J.},
  journal={IEEE Transactions on Robotics},
  volume={32},
  number={6},
  pages={1309--1332},
  year={2016}
}

@article{mur_artal_2017_orbslam2,
  title={ORB-SLAM2: An Open-Source SLAM System for Monocular, Stereo, and RGB-D Cameras},
  author={Mur-Artal, Raul and Tard{\'o}s, Juan D.},
  journal={IEEE Transactions on Robotics},
  volume={33},
  number={5},
  pages={1255--1262},
  year={2017}
}

@inproceedings{geiger2012kitti,
  title={Are we ready for Autonomous Driving? The KITTI Vision Benchmark Suite},
  author={Geiger, Andreas and Lenz, Philip and Urtasun, Raquel},
  booktitle={Proceedings of the IEEE/CVF Conference on Computer Vision and Pattern Recognition},
  pages={2443--2451},
  year={2012}
}

@inproceedings{sun2020waymo,
  title={Scalability in Perception for Autonomous Driving: Waymo Open Dataset},
  author={Sun, Pei and Kretzschmar, Henrik and Dotiwalla, Xerxes and Chouard, Aurelien and Casas, Sergio and Lin, Wenjie and Sadat, Abbas and Varadarajan, Balakrishnan and Shlens, Jonathon and Chen, Zhifeng and Yuille, Alan and Anguelov, Dragomir},
  booktitle={Proceedings of the IEEE/CVF Conference on Computer Vision and Pattern Recognition},
pages={2443--2451},
  year={2020}
}

@inproceedings{klein2007ptam,
  title={Parallel Tracking and Mapping for Small AR Workspaces},
  author={Klein, Georg and Murray, David},
  booktitle={IEEE and ACM International Symposium on Mixed and Augmented Reality},
pages={225--234},
  year={2007}
}

@inproceedings{izadi2011kinectfusion,
  title={KinectFusion: Real-time 3D Reconstruction and Interaction Using a Moving Depth Camera},
  author={Izadi, Shahram and Kim, David and Hilliges, Otmar and Molyneaux, David and Newcombe, Richard and Kohli, Pushmeet and Shotton, Jamie and Hodges, Steve and Freeman, Daniel and Davison, Andrew and Fitzgibbon, Andrew},
  booktitle={ACM Symposium on User Interface Software and Technology (UIST)},
  pages={559--568},
  year={2011}
}

@inproceedings{choy2019minkowski,
  title={4D Spatio-Temporal ConvNets: Minkowski Convolutional Neural Networks},
  author={Choy, Christopher and Gwak, JunYoung and Savarese, Silvio},
  booktitle={Proceedings of the IEEE/CVF Conference on Computer Vision and Pattern Recognition},
  pages={3075--3084},
  year={2019}
}

@inproceedings{thomas2019kpconv,
  title={KPConv: Flexible and Deformable Convolution for Point Clouds},
  author={Thomas, Hugues and Qi, Charles R. and Deschaud, Jean-Emmanuel and Marcotegui, Beatriz and Goulette, Fran{\c{c}}ois and Guibas, Leonidas J.},
  booktitle={Proceedings of the IEEE/CVF Conference on Computer Vision and Pattern Recognition},
  pages={6410--6419},
  year={2019}
}

@inproceedings{jia2021align,
  title={Scaling Up Visual and Vision-Language Representation Learning with Noisy Text Supervision},
  author={Jia, Chao and Yang, Yinfei and Xia, Ye and Chen, Yi-Ting and Parekh, Zarana and Pham, Hieu and Le, Quoc V. and Sung, Yunhsuan and Li, Zhen and Duerig, Thomas},
  booktitle={International Conference on Machine Learning (ICML)},
  pages={4904--4916},
  year={2021}
}

@article{vild2021,
  title={Open-Vocabulary Object Detection via Vision and Language Knowledge Distillation},
  author={Gu, Xiuye and Kuo, Yen-Chun and Cui, Yin and Sun, Zecheng and Zhang, David and Hoi, Steven C. H.},
  journal={arXiv preprint arXiv:2104.13921},
  year={2021}
}

@inproceedings{detic2022,
  title={Detecting Twenty-thousand Classes using Image-level Supervision},
  author={Zhou, Xingyi and Girdhar, Rohit and Joulin, Armand and Kr{\"a}henb{\"u}hl, Philipp and Misra, Ishan},
  booktitle={European Conference on Computer Vision},
  pages={350--368},
  year={2022}
}

@inproceedings{openscene2023,
  title={OpenScene: 3D Scene Understanding with Open Vocabularies},
  author={Peng, Songyou and Genova, Kyle and Jiang, Chiyu Max and Tagliasacchi, Andrea and Pollefeys, Marc and Funkhouser, Thomas and Nie{\ss}ner, Matthias and Liu, Sida Peng},
  booktitle={Proceedings of the IEEE/CVF Conference on Computer Vision and Pattern Recognition},
pages={1786--1796},
  year={2023}
}

@article{tian2025ccl,
  title={CCL-LGS: Contrastive Codebook Learning for 3D Language Gaussian Splatting},
  author={Tian, Lei and Li, Xiaomin and Ma, Liqian and Huang, Hefei and Zheng, Zirui and Yin, Hao and Li, Taiqing and Lu, Huchuan and Jia, Xu},
  journal={arXiv preprint arXiv:2505.20469},
  year={2025}
}

@InProceedings{Sun_2020_CVPR,
  author    = {Sun, Pei and Kretzschmar, Henrik and Dotiwalla, Xerxes and Chouard, Aurelien and Patnaik, Vijaysai and Tsui, Paul and Guo, James and Zhou, Yin and Chai, Yuning and Caine, Benjamin and Vasudevan, Vijay and Han, Wei and Ngiam, Jiquan and Zhao, Hang and Timofeev, Aleksei and Ettinger, Scott and Krivokon, Maxim and Gao, Amy and Joshi, Aditya and Zhang, Yu and Shlens, Jonathon and Chen, Zhifeng and Anguelov, Dragomir},
  title     = {Scalability in Perception for Autonomous Driving: Waymo Open Dataset},
  booktitle = {Proceedings of the IEEE/CVF Conference on Computer Vision and Pattern Recognition (CVPR)},
  month     = {June},
  year      = {2020}
}

@article{eckhardt1980weber,
  title={Weber's problem and Weiszfeld's algorithm in general spaces},
  author={Eckhardt, Ulrich},
  journal={Mathematical Programming},
  volume={18},
  number={1},
  pages={186--196},
  year={1980},
  publisher={Springer}
}
}

\clearpage
\setcounter{page}{1}
\maketitlesupplementary
\appendix

In this supplementary material, we provide additional details omitted from the main manuscript. Sec.~\ref{subsec:impl} describes the implementation details and the 3D tasks under evaluation. Sec.~\ref {subsec:eval_prot} outlines the experimental setup and the 3D semantic segmentation evaluation protocol on 3D Gaussian Splatting. Sec.~\ref{subsec:robust} further presents a robustness study, where we stress-test our method under corrupted SAM masks to assess performance degradation in noisy segmentation scenarios.  while Sec.~\ref{subsec:add_results} presents qualitative results, annotation analyses, and city-scale evaluations. Finally, Sec.~\ref{subsec:limt} discusses limitations and future directions. 

\section{Implementation Details}
\label{subsec:impl}

Our method operates in two stages. In the pre-training stage, we apply the ViT-H variant of SAM~\cite{kirillov2023sam} to segment each image. Multi-level language feature maps are then extracted with OpenCLIP ViT-B/16~\cite{radford2021clip}, from which we derive per-patch language embeddings. In parallel, we optimize the 3D Gaussian Splatting parameters~\cite{kerbl20233d} using the standard training pipeline with the \emph{gsplat} rasterizer~\cite{ye2025gsplat}, running 30k iterations. Unlike the original rasterizer, \emph{gsplat} natively supports rendering high-dimensional Gaussian attributes, which enables evaluation on 2D open-vocabulary tasks.

In the subsequent forward-rendering stage, we adopt the feature aggregation strategy of Occam’s LGS~\cite{cheng2024occam}. For each Gaussian within the view frustum, we compute its center-projected pixel location and extract the corresponding 2D language feature $f_i^s$. Simultaneously, we record its marginal contribution $w_i(r)$ as defined in Eq.~\eqref{eq:visibility_weights}, and retain the most visible Gaussians following the gating strategy in Sec.~\ref{sec:gating}. The selected Gaussians are then robustly aligned with multi-view features through our streaming aggregation in cosine space, described in Sec.~\ref{sec:median}.

This entire process completes within 10 seconds to one minute per scene (depending on scene scale) without memory overflow. All experiments are conducted on an NVIDIA RTX 4090 GPU.

\section{Evaluation Protocols}
\label{subsec:eval_prot}
We only compare results following the same evaluation protocol and re-evaluate those prior works that followed other protocols.

\textbf{Datasets}
We evaluate our method on two datasets: LERF-OVS~\cite{qin2024langsplat} and ScanNet~\cite{dai2017scannet}. LERF-OVS consists of four scenes (teatime, waldo\_kitchen, figurines, ramen), each annotated with pixel-wise semantic masks and paired with short text queries. In this dataset, we evaluate open-vocabulary object selection in both 2D and 3D settings. To further evaluate 3D semantic segmentation, we adopt a Gaussian-based evaluation protocol on ScanNet, a large-scale RGB-D dataset for indoor scene understanding. Each ScanNet sequence is reconstructed into a textured 3D mesh with globally aligned camera poses and semantic annotations. We select eight representative scenes covering diverse indoor environments, including living rooms, bathrooms, kitchens, bedrooms, and meeting rooms.

\textbf{2D and 3D Evaluation on the LERF-OVS Dataset.}
For the 2D evaluation, we follow the protocol of LERF~\cite{kerr2023lerf}: 512-dimensional feature maps are rendered, and a relevancy map is computed with respect to the CLIP-embedded text query. The relevancy map is then thresholded at 0.5 to obtain the predicted binary mask. For the 3D evaluation, we adopt the protocol of OpenGaussian~\cite{wu2024opengaussian}, where the relevancy score between each 3D Gaussian’s language embedding and the text query embedding is computed and thresholded at 0.6. The alpha values of the selected Gaussians are subsequently projected onto the image plane to generate the predicted mask. In both cases, the predicted masks are compared against the GT annotations of the LERF-OVS dataset.

\textbf{3D Semantic Segmentation on the ScanNet-v2 Dataset.}
Previous works on 3D semantic segmentation~\cite{wu2024opengaussian, li2024instancegaussian} typically freeze the input point cloud (derived from ground-truth annotations) during 3D Gaussian Splatting training to cope with the absence of GT labels as the point clouds evolve. However, this strategy degrades the 2D rendering quality of 3DGS. We instead propagate ground-truth (GT) labels from the annotated point cloud to the Gaussians, thereby obtaining pseudo-GT labels at each Gaussian’s 3D mean. Following OpenGaussian~\cite{wu2024opengaussian}, we evaluate on subsets of 19, 15, and 10 of the 40 most common classes. For each class, we encode the text label using CLIP~\cite{radford2021clip} to obtain a 512-dimensional embedding, and compute its cosine similarity with the registered language features of each Gaussian. Each Gaussian is then assigned to the class with the highest similarity score. Performance is measured in terms of mIoU and mAcc against the pseudo-GT Gaussian point cloud.

\noindent\textbf{Pseudo-Gaussian Labeling.}
% Inspired by Kim~\etal~\cite{drsplat25}, we instead first train 3D Gaussian Splatting without restrictions and subsequently propagate labels from the ground-truth point cloud to the optimized Gaussians, explicitly accounting for their scales and rotations. 
Given optimized Gaussians 
$\Theta=\{\theta_i\}_{i=1}^N$ 
with center $\mu_i$, scale $s_i=(s_{ix},s_{iy},s_{iz})$, rotation $R_i$ 
(hence $\Sigma_i = R_i \,\mathrm{diag}(s_i^2)\,R_i^\top$), and opacity $\alpha_i$, 
and a labeled point cloud $\{(p_k,s_{p_k})\}_{k=1}^Q$, 
we assign a semantic label to each Gaussian by respecting the \emph{true} 3DGS geometry 
and the compositing kernel. 
In contrast to prior protocols, which (i) maximize the \emph{sum of Mahalanobis distances} over class points to assign a single label, and (ii) require dense all-pairs computations, our approach assigns semantic labels by respecting the \emph{true} 3DGS geometry and properties.
Specifically, we evaluate the density contribution of a point $p$ to the Gaussian $\mu_i$:
\begin{equation}
w_i(p) = \exp\!\left(-\tfrac{1}{2}\, d_i^2(p)\right),
\end{equation}
where $d_i^2(p)$ denotes the squared Mahalanobis distance.

Since boundary Gaussians may be partially transparent or occupy negligible volume, 
we further modulate the votes with a per-Gaussian significance term:
\begin{equation}
\gamma_i = \alpha_i \, s_{ix} s_{iy} s_{iz}, 
\quad\quad
w_i(p) \leftarrow \gamma_i \, w_i(p).
\end{equation}
This ensures consistency with the volume-aware IoU metric, which weights Gaussians by both opacity and ellipsoid volume.

\begin{figure*}[t]
    \centering
    \includegraphics[width=\textwidth]{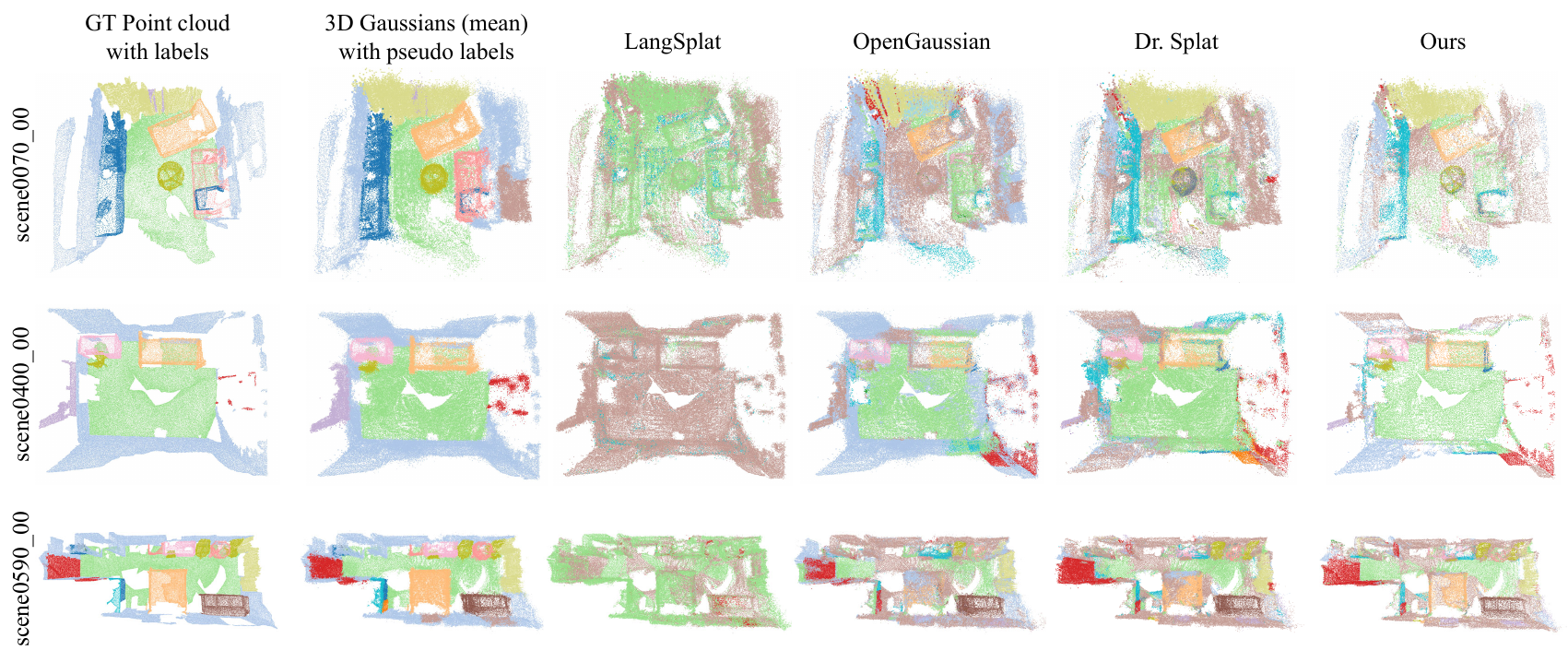}
    \caption{More qualitative results of 3D semantic segmentation on the ScanNet-v2 dataset~\cite{dai2017scannet},}
    \label{fig:more_qualitative_scannet}
\end{figure*}

Finally, instead of constructing an $N\times Q$ all-pairs distance matrix, we build a per-Gaussian candidate set $K_i$ via spatial culling with an adaptive radius
\[
radius_i = \tau \cdot \max(s_i),
\]
with a top-$k$ fallback to handle sparse neighborhoods. 
We then compute $d_i^2(\cdot)$ only for $p_k \in K_i$, 
processing Gaussians in GPU-friendly chunks. 
This reduces the complexity from $O(NQ)$ to $O(\sum_i |K_i|)$ 
and the memory footprint from $O(NQ)$ to $O(|K|)$, 
while retaining only geometrically plausible candidates under each anisotropic ellipsoid. The generated Gaussian point clouds with pseudo GT labels are illustrated in \figureautorefname~\ref{fig:qualitative_scannet} and \figureautorefname~\ref{fig:more_qualitative_scannet} (the second column from left to right). 

\section{Robustness Evaluation with Perturbed Masks}
\label{subsec:robust}
\begin{figure}[tbp]
  \centering
  \includegraphics[width=0.95\linewidth]{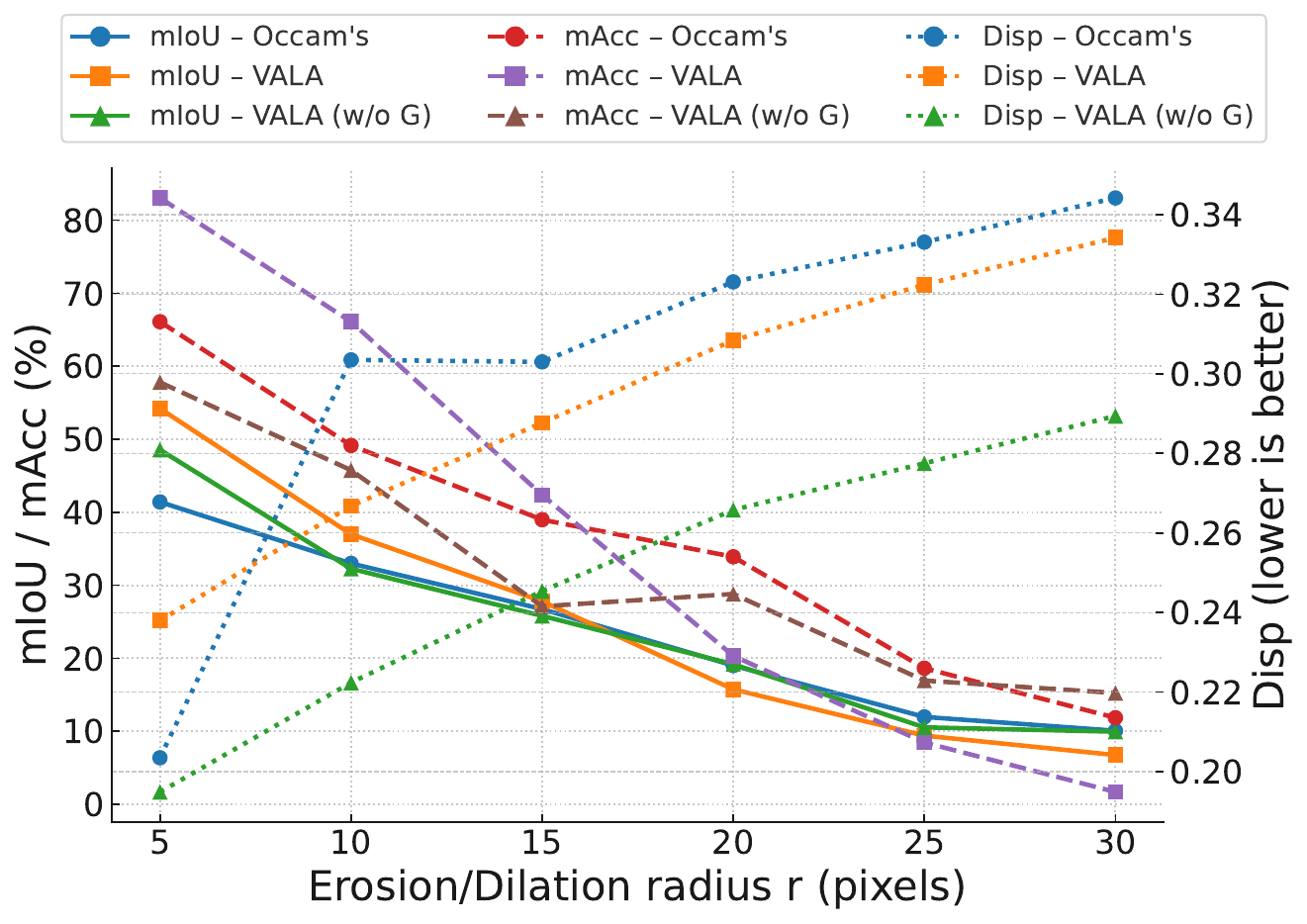} % png/jpg/pdf 都行
  \caption{Robustness under mask boundary corruptions. mIoU/mAcc (\%) are shown on the left $y$-axis; \textit{Disp} (lower is better) on the right $y$-axis. 
  We vary the erosion/dilation radius $r$ (pixels). 
  VALA degrades more slowly than Occam’s and its ablation without gating (VALA w/o G), while achieving lower \textit{Disp} across severities.}
  \label{fig:robustness}
\end{figure}

To evaluate robustness against segmentation noise, we perform an experiment on the teatime scene of LERF-OVS by simulating errors in SAM masks.

\textbf{Stress-Testing Robustness with Corrupted Masks.}
To stress-test robustness against imperfect proposals, 
we corrupt each SAM mask by a per-mask morphological perturbation applied at the original image resolution. 
Let $m_k \in \{0,1\}^{H\times W}$ denote the binary mask of instance $k$, 
and let
\[
B_r = \{(x,y)\in\mathbb{Z}^2 : x^2 + y^2 \leq r\}
\]
be a disk-shaped structuring element of radius $r$ pixels, where $r \in {5,10,15,20,25,30}$, to simulate different perturbation levels.

For every mask we draw an independent sign variable $\sigma_k \in \{-1,+1\}$ with equal probability 
$P(\sigma_k=+1)=P(\sigma_k=-1)=0.5$. 
The corrupted mask $\tilde{m}_k$ is then
\[
\tilde{m}_k =
\begin{cases}
m_k \,\ominus\, B_r, & \text{if } \sigma_k=-1 \quad \text{(erosion)}, \\[4pt]
m_k \,\oplus\, B_r, & \text{if } \sigma_k=+1 \quad \text{(dilation)},
\end{cases}
\]
where $\ominus$ and $\oplus$ denote morphological erosion and dilation, respectively.  

To prevent degenerate outcomes on small objects, we enforce a non-vanishing guard: 
if erosion yields an empty or tiny region (area below a minimum threshold $\tau_{\min}$ pixels), 
we fallback to dilation and set $\tilde{m}_k \leftarrow m_k \oplus B_r$.  
After corruption, we recompute tight bounding boxes from $\tilde{m}_k$ 
and propagate them to downstream steps (e.g., cropping and $224\times224$ resizing for CLIP feature extraction).  

This perturbation stochastically shifts boundaries outward/inward by approximately $r$ pixels 
while preserving instance identity, thereby simulating over- and under-segmentation errors 
commonly observed in practice.

\textbf{Evaluation Protocal.}
To assess the robustness of the proposed streaming median in the cosine space, we compare three variants: the baseline Occam's LGS~\cite{cheng2024occam}, our full model incorporating both visibility-aware gating and robust multi-view aggregation (VALA), and an ablation variant with only the robust multi-view aggregation module (VALA w/o G). In addition to the standard mIoU and mAcc metrics for evaluating the final 3D object selection task, we further introduce the \emph{dispersion} score, which specifically quantifies the robustness of assigned language features under multi-view variations.
Given a Gaussian $g_i$ with observed unit features $f_{i}^s \in \mathbb{S}^{d-1}$, the per-Gaussian dispersion is computed as
\begin{equation}
\mathrm{Disp}_i = \frac{1}{|S_i|}\sum_{(i,s)\in S_i}\Big(1-\langle f_i^s, z_i^*\rangle\Big),
\end{equation}

\noindent
At the scene level, we report the average:

\begin{equation}
\mathrm{Disp}_{\text{scene}} = \frac{1}{|I|}\sum_{i\in I}\mathrm{Disp}_i,
\end{equation}

\noindent
This metric captures the average misalignment between observed features and the aggregated Gaussian feature, where lower values indicate higher consistency.

\begin{figure*}[t]
    \centering
    \includegraphics[width=\textwidth]{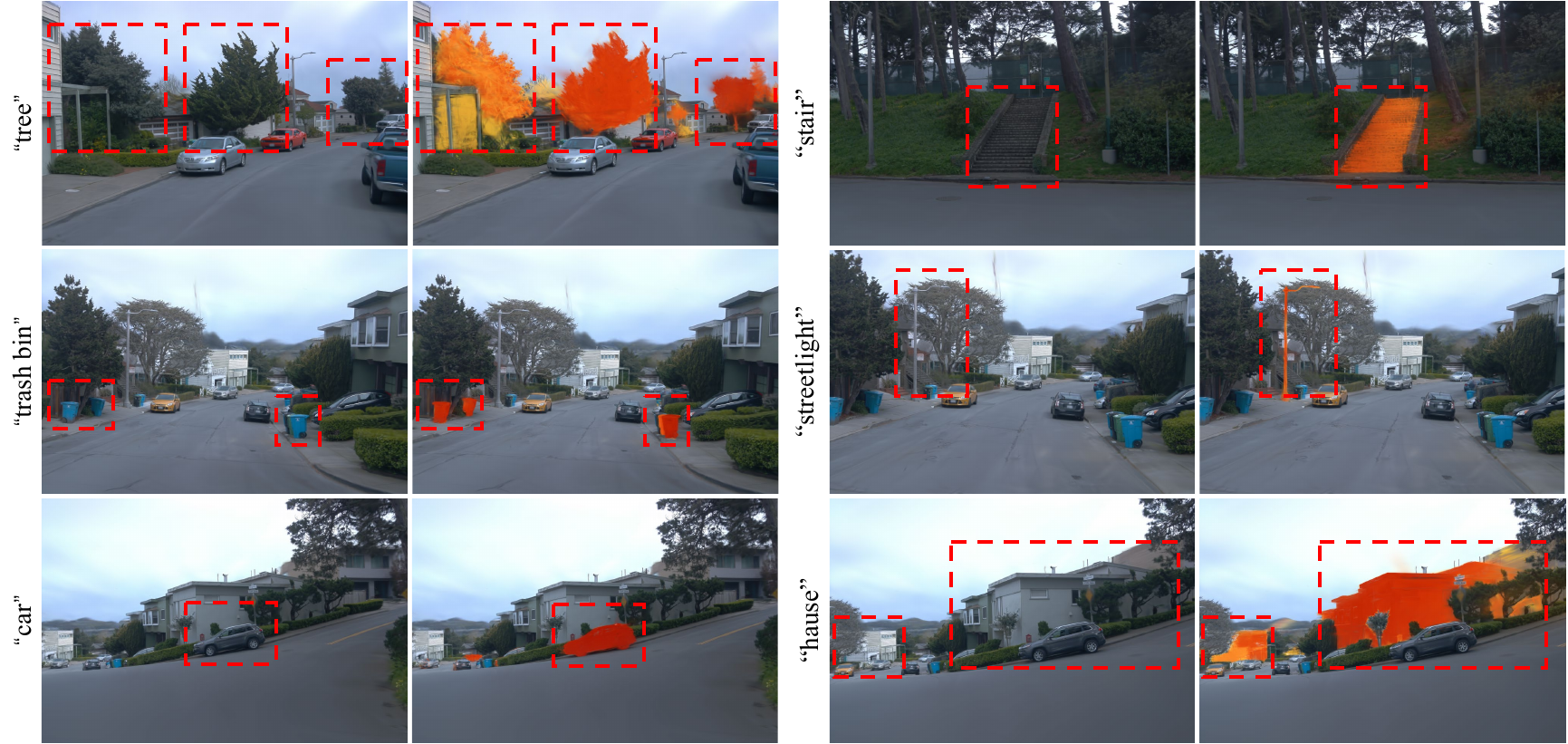}
    \caption{\textbf{Qualitative results on the Waymo Open Dataset~\cite{Sun_2020_CVPR}}. The colored regions indicate the activation maps corresponding to the given text prompts.}
    \label{fig:quantitative_waymo}
\end{figure*}

\textbf{Results Analysis.}
The results are presented in \figureautorefname~\ref{fig:robustness}. As the corruption radius increases from $r=5$ to $30$ px, all methods show a monotonic decline in mIoU/mAcc and a corresponding rise in Disp, confirming that boundary noise simultaneously degrades semantic accuracy and cross-view consistency. Importantly, the deterioration is substantially slower for our methods than for Occam’s LGS, as reflected by the smaller slope of Disp. In terms of accuracy, VALA achieves the strongest results: at $r=5$, it surpasses Occam’s by +12.8 mIoU and +17.0 mAcc, with substantial gains still observed at $r=10$. Meanwhile, the Disp values reveal a complementary trend—although VALA’s Disp is marginally higher than Occam’s at $r=5$, it drops below Occam’s from $r=10$ onwards. This demonstrates that the combination of visibility-aware gating and robust aggregation not only improves accuracy but also enhances multi-view consistency in the practically relevant regime of mild mask noise.

When boundary damage becomes severe, however, the picture changes. VALA (w/o G) overtakes the full VALA model in accuracy (e.g., at $r=30$, achieving 9.95/15.25 vs.\ 6.75/1.69 in mIoU/mAcc) and consistently yields the lowest Disp across all radii. This suggests that the fixed gating threshold becomes overly conservative under extreme corruption, discarding too many observations and leaving insufficient evidence for many Gaussians. In contrast, the cosine-median aggregator alone remains robust, preserving both accuracy and consistency in this challenging setting. Overall, these results highlight a clear regime split: visibility-aware gating combined with a cosine median provides the strongest accuracy and consistency under realistic (mild to moderate) noise. However, under extreme boundary corruption, robust aggregation is the key factor, as overly strict gating thresholds reduce coverage and performance.

\section{Additional Results}
\label{subsec:add_results}
In this section, we present additional results on the ScanNet dataset and, more importantly, demonstrate that our algorithm can be applied to real-world outdoor datasets, achieving superior open-vocabulary semantic segmentation in autonomous driving scenarios.

\textbf{More Qualitative Results on the ScanNet Dataset.}
We provide additional qualitative results on three bedrooms with varying levels of complexity and clutter. Across all scenes, competing methods struggle to correctly recognize the bed (highlighted in orange); the occluded portions near the wall are consistently misclassified as adjacent categories, such as the wall or floor. This issue persists in the third scene, where the bed is fragmented into multiple categories. In contrast, our method preserves the bed as a coherent instance, owing to the proposed gating module that explicitly handles low-visibility Gaussians.

\textbf{Experiments on the Waymo Open Dataset.} To further validate our algorithm’s generalization capability in real-world outdoor environments, we conduct experiments on the Waymo Open Dataset~\cite{Sun_2020_CVPR}. This dataset is a large-scale, high-quality autonomous driving benchmark that provides synchronized LiDAR and multi-camera data collected across diverse urban and suburban geographies, along with comprehensive 2D/3D annotations and tracking identifiers.
For evaluation, we select a sequence captured in a residential neighborhood that contains rich semantic elements, such as vehicles, vegetation, street infrastructure, and buildings. We focus on five of the most common outdoor categories, e.g. \textit{tree}, \textit{trash bin}, \textit{car}, \textit{streetlight}, and \textit{house}, as well as one tail category, \textit{stair}. The qualitative results in \figureautorefname~\ref{fig:quantitative_waymo} demonstrate that our method achieves precise open-vocabulary 3D semantic segmentation on outdoor data. Both small-scale objects (e.g., trash bins and streetlights) and large-scale objects (e.g., trees, cars, and houses) are not only correctly retrieved but also segmented with sharp boundaries, reflecting the accurate registration of language features on the 3D Gaussian Splatting representation. Notably, our method remains robust under occlusion owing to the proposed visibility-aware gating module. For example, correctly delineating trees behind metallic structures or houses partially obscured by vegetation.

These findings emphasize the robustness and versatility of our method when transferred from indoor (ScanNet) to challenging outdoor driving scenarios, underscoring its strong potential for real-world autonomous driving applications. A supplementary video is included to further demonstrate the effectiveness and the multi-view consistency of our method. 

\section{Limitations}
\label{subsec:limt}

While our approach demonstrates strong performance across multiple tasks, including 2D and 3D object selection as well as 3D semantic segmentation, and exhibits notable generalization to cross-domain settings such as outdoor datasets, certain limitations remain. To assess robustness against noisy SAM masks, we conducted stress tests with multi-scale morphological perturbations. The results show that our visibility-aware gating achieves superior mIoU and mAcc under moderate noise, while the proposed cosine median maintains low dispersion even under severe corruption, indicating the effectiveness of our robust feature aggregator. However, our current framework relies on a fixed threshold to prune Gaussians, which can become overly conservative under extreme noise, resulting in degraded multi-view consistency. Moreover, our method is specifically designed for static scenes and does not naturally extend to dynamic environments. Future work will therefore focus on developing adaptive, scene-aware thresholds and extending our framework to handle dynamic scenes.

\end{document}